\documentclass[sigconf]{acmart}

\copyrightyear{2025}
\acmYear{2025}
\setcopyright{acmlicensed}
\acmConference[KDD '25] {Proceedings of the 31st ACM SIGKDD Conference on Knowledge Discovery and Data Mining V.2}{August 3--7, 2025}{Toronto, ON, Canada.}
\acmBooktitle{Proceedings of the 31st ACM SIGKDD Conference on Knowledge Discovery and Data Mining V.2 (KDD '25), August 3--7, 2025, Toronto, ON, Canada}
\acmISBN{979-8-4007-1454-2/25/08}
\acmDOI{10.1145/3711896.3736899}

\usepackage{marvosym}
\usepackage{multirow}
\usepackage{subfig}
\usepackage{caption}

\begin{document}

\title{CrossLinear: Plug-and-Play Cross-Correlation Embedding for Time Series Forecasting with Exogenous Variables}

\author{Pengfei Zhou}
\orcid{0009-0007-6369-3181}
\affiliation{
  \institution{University of Science and Technology of China}
  \city{Hefei}
  \country{China}
}
\email{pengfeizhou@mail.ustc.edu.cn}

\author{Yunlong Liu}
\orcid{0009-0009-9101-3942}
\affiliation{
  \institution{University of Science and Technology of China}
  \city{Hefei}
  \country{China}
}
\email{liuyunlong@mail.ustc.edu.cn}

\author{Junli Liang}
\orcid{0009-0001-6112-7908}
\affiliation{
  \institution{University of Science and Technology of China}
  \city{Hefei}
  \country{China}
}
\email{jlliang@mail.ustc.edu.cn}

\author{Qi Song \Letter}
\orcid{0000-0002-1726-7858}
\affiliation{
  \institution{University of Science and Technology of China}
  \city{Hefei}
  \country{China}
}
\affiliation{
  \institution{Deqing Alpha Innovation Institute}
  \city{Huzhou}
  \country{China}
}
\email{qisong09@ustc.edu.cn}

\author{Xiangyang Li}
\orcid{0000-0002-6070-6625}
\affiliation{
  \institution{University of Science and Technology of China}
  \city{Hefei}
  \country{China}
}
\affiliation{
  \institution{Deqing Alpha Innovation Institute}
  \city{Huzhou}
  \country{China}
}
\email{xiangyangli@ustc.edu.cn}

\begin{abstract}
Time series forecasting with exogenous variables is a critical emerging paradigm that presents unique challenges in modeling dependencies between variables. Traditional models often struggle to differentiate between endogenous and exogenous variables, leading to inefficiencies and overfitting. In this paper, we introduce \textbf{CrossLinear}, a novel Linear-based forecasting model that addresses these challenges by incorporating a plug-and-play cross-correlation embedding module. This lightweight module captures the dependencies between variables with minimal computational cost and seamlessly integrates into existing neural networks. Specifically, it captures time-invariant and direct variable dependencies while disregarding time-varying or indirect dependencies, thereby mitigating the risk of overfitting in dependency modeling and contributing to consistent performance improvements. Furthermore, CrossLinear employs patch-wise processing and a global linear head to effectively capture both short-term and long-term temporal dependencies, further improving its forecasting precision. Extensive experiments on 12 real-world datasets demonstrate that CrossLinear achieves superior performance in both short-term and long-term forecasting tasks. The ablation study underscores the effectiveness of the cross-correlation embedding module. Additionally, the generalizability of this module makes it a valuable plug-in for various forecasting tasks across different domains. Codes are available at \href{https://github.com/mumiao2000/CrossLinear}{https://github.com/mumiao2000/CrossLinear}.
\end{abstract}




\maketitle

\newcommand\kddavailabilityurl{https://doi.org/10.5281/zenodo.15482271}

\ifdefempty{\kddavailabilityurl}{}{
\begingroup\small\noindent\raggedright\textbf{KDD Availability Link:}\\
The source code of this paper has been made publicly available at \url{\kddavailabilityurl}.
\endgroup
}

\section{Introduction}
\begin{figure}[!tbp]
\centering
\includegraphics[width=0.825\linewidth]{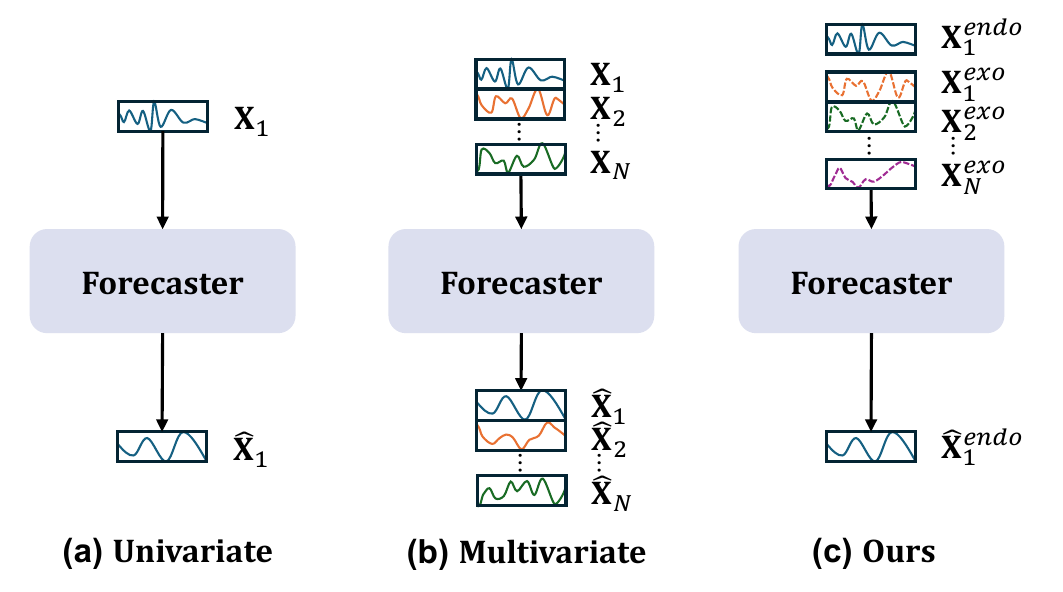}
\caption{Different forecasting paradigms. Here, \(\mathbf{X}_{i}\) represents the \(i^\mathrm{th}\) variable, \(\mathbf{X}^{endo}_{1}\) is the endogenous variable, and \(\mathbf{X}^{exo}_{i}\) denotes the \(i^\mathrm{th}\) exogenous variable. (a) Univariate forecasting. (b) Multivariate forecasting. (c) Forecasting with exogenous variables.}
\label{fig: Paradigm}
\end{figure}

Time series forecasting is extensively applied in various fields such as finance~\cite{finance_1,finance_2}, meteorology~\cite{meteorology_1}, medicine~\cite{medicine_1}, transportation~\cite{transportation_1,transportation_2}, and industrial control~\cite{industrial_1}. This task is traditionally categorized into two paradigms: \textbf{univariate forecasting} and \textbf{multivariate forecasting}~\cite{univaraite_multivariate_1,univaraite_multivariate_2,univaraite_multivariate_3}, corresponding to one-to-one and many-to-many tasks, as illustrated in Figures \ref{fig: Paradigm}(a) and \ref{fig: Paradigm}(b). However, in real-world applications, not all variables are targets of forecasting. \textbf{Forecasting with exogenous variables}~\cite{timexer,exogenous_1,exogenous_2,nbeatsx}, \textit{i.e.}, many-to-one tasks, represents a critical emerging paradigm (Figure \ref{fig: Paradigm}(c)). For instance, traffic volume is often influenced by various exogenous variables~\cite{traffic_volume}. Holidays, for example, can lead to increased highway traffic due to travel while reducing urban commuting. Weather conditions also play a crucial role. Adverse weather like heavy rain, snow, or fog slows down vehicles and increases travel time, whereas clear weather promotes smoother traffic flow. A one-to-one model relying solely on historical traffic data fails to capture these exogenous influences, while many-to-many models introduce inefficiencies by forecasting unnecessary exogenous variables. Understanding dependencies between endogenous and exogenous factors can enhance forecasting performance.

Compared with traditional paradigms, the many-to-one time series forecasting paradigm introduces new challenges in modeling the dependencies between endogenous and exogenous variables. Since the focus is on the endogenous variable, the model must effectively capture the effects of exogenous variables. Existing methods generally address variable dependencies through two approaches: channel (\textit{aka.} variable) independent (CI) and channel dependent (CD). CI methods process each variable separately in the neural network, avoiding explicit variable dependency modeling but relying on embedding layers to implicitly capture such relationships~\cite{autoformer,dlinear,patchtst,rlinear,timesnet}. Conversely, CD methods, such as those employing cross-attention or graph convolution, explicitly model variable dependencies~\cite{crossformer,itransformer,msgnet,moderntcn}. However, these approaches often fail to differentiate endogenous and exogenous variables, resulting in unnecessary interactions among exogenous variables, which increases model complexity.

An intuitive assumption is that CD methods should outperform CI methods, as they explicitly model variable dependencies. However, numerous studies have demonstrated that CI methods often achieve better results~\cite{patchtst,dlinear,timesnet,sparsetsf}. A key reason for this paradox lies in the inherent complexity of variable dependencies, which makes them challenging to capture accurately. For instance, exogenous variables may exhibit positive correlations with the endogenous variable during certain periods but negative correlations at other times~\cite{msgnet}. As a result, deep models are prone to being influenced by irrelevant or uninformative features. Therefore, a crucial challenge in the many-to-one forecasting paradigm is to integrate exogenous variables without degrading model performance due to erroneous dependency modeling.

Beyond modeling variable dependencies, it is also crucial to account for temporal dependencies in the many-to-one paradigm~\cite{long_dataset,fcstgnn}. The model's ability to capture both types of dependencies is key to its representational power. Transformer-based models have shown strong performance in learning these dependencies~\cite{itransformer}. However, existing models, such as Crossformer~\cite{crossformer}, are not specifically designed for the many-to-one paradigm and fail to differentiate between endogenous and exogenous variables. TimeXer~\cite{timexer}, a model tailored for this forecasting paradigm, introduces additional cross-attention mechanisms to address this issue. Nevertheless, Transformer-based models often suffer from high computational complexity and resource demands due to the attention mechanism, making them impractical for real-world application. In contrast, some Linear-based models like DLinear~\cite{dlinear} and RLinear~\cite{rlinear} offer lower complexity but rely on CI approaches, which fail to adequately capture variable dependencies. While TiDE~\cite{tide} explicitly incorporates exogenous variable information, its actual performance under the many-to-one paradigm is not very good. In addition to Transformer- and Linear-based models, there are also CNN-based models~\cite{timesnet,moderntcn} and GNN-based models~\cite{fcstgnn,msgnet}. However, these models also face similar challenges in adapting to the many-to-one forecasting paradigm.

Inspired by position embedding in Transformers~\cite{transformer}, we use convolutional operations to generate cross-correlation embeddings that capture the time-invariant and direct dependencies between endogenous and exogenous variables. This embedding method can be seamlessly integrated into existing networks as a lightweight, nearly cost-free module. Building on this concept, we introduce \textbf{CrossLinear}. Specifically, CrossLinear leverages the cross-correlation embedding module to model the dependencies between endogenous and exogenous variables and incorporates patch embedding for temporal dependency modeling. Through cross-correlation embedding, CrossLinear focuses on the endogenous variable while integrating information from exogenous variables, thereby ensuring performance improvement. Furthermore, the inclusion of patch embedding and global forecasting head enables the model to capture short-term and long-term temporal dependencies.

Extensive experiments demonstrate that CrossLinear achieves excellent performance on various real-world datasets. Our key contributions are summarized as follows:

\noindent\textbf{(1) Plug-and-play cross-correlation embedding module}: We propose a lightweight cross-correlation embedding module that integrates information from exogenous variables with the endogenous variable at minimal computational cost. This module can be seamlessly incorporated into other models as a plug-in, enabling CI models to effectively capture variable dependencies.

\noindent\textbf{(2) Patch-wise processing}: Patch-wise processing is adapted in CrossLinear to capture both short-term and long-term temporal dependencies. This approach not only enhances the model’s ability to handle temporal dependencies but also improves robustness in scenarios involving missing data.

\noindent\textbf{(3) Versatility across various paradigms}: Extensive experiments show that CrossLinear excels in many-to-one tasks and strikes an optimal balance between
training time and error. Furthermore, by employing weight-sharing mechanism, the proposed model can be effectively extended to many-to-many tasks, \textit{i.e.}, multivariate forecasting, demonstrating its broad applicability.

\section{Related Work}
We categorize the related work as follows.

\begin{figure*}[!htbp]
\centering
\includegraphics[width=0.85\linewidth]{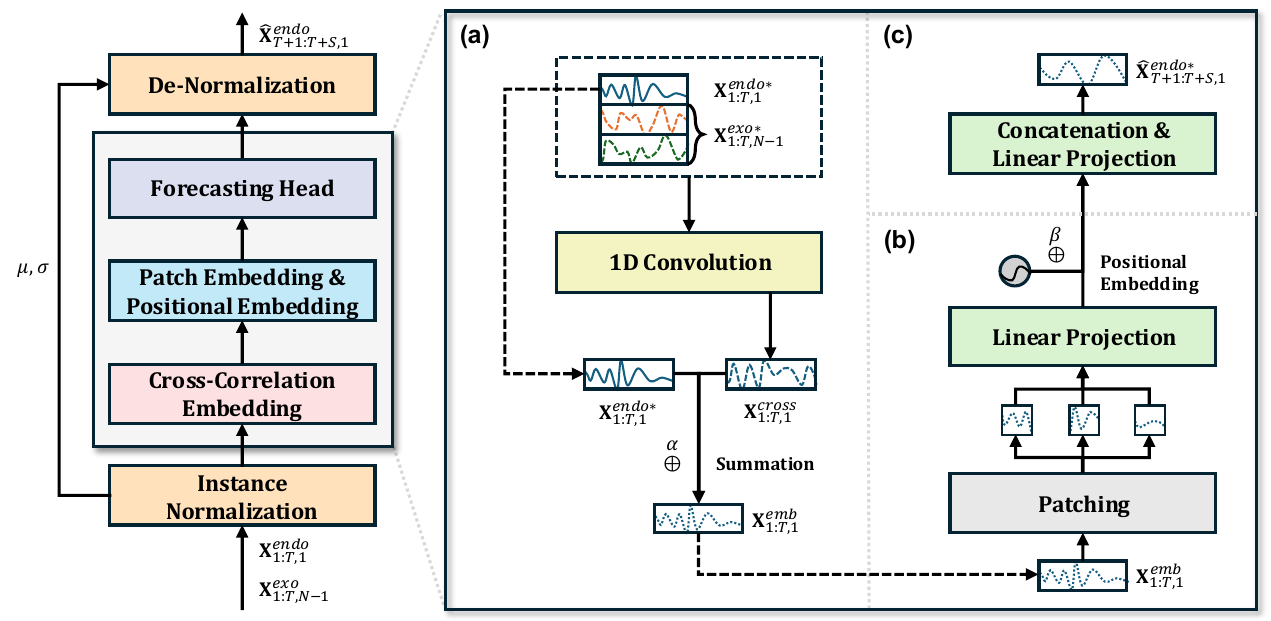}
\caption{Architecture of CrossLinear. (a) The cross-correlation embedding module captures variable dependencies. (b) The patch embedding module captures short-term temporal dependencies. (c) The linear head is responsible for capturing long-term patterns.}
\label{fig: Model Architecture}
\end{figure*}

\subsection{Channel Independent or Dependent?}
In multivariate time series forecasting, where variables exhibit interdependencies, some methods adopt a CD strategy~\cite{crossformer,itransformer,msgnet,tide,timexer}. Transformer-based models, such as Crossformer~\cite{crossformer} and iTransformer~\cite{itransformer}, capture both temporal and variable dependencies using mechanisms like self-attention and cross-attention. Linear-based models, including TSMixer~\cite{tsmixer}, inspired by MLP-Mixer~\cite{mlpmixer}, transpose multivariate time series to capture dependencies, while GNNs, such as FourierGNN~\cite{fouriergnn}, leverage hypervariate graphs for comprehensive modeling.

Surprisingly, some CI methods outperform CD ones. PatchTST~\cite{patchtst} enhances efficiency via patching, and even simple Linear-based models like DLinear~\cite{dlinear} and RLinear~\cite{rlinear} achieve competitive results. Models relying on time-frequency decomposition, such as Autoformer~\cite{autoformer} and TimesNet~\cite{timesnet}, face challenges in CD settings due to varying variable periodicities.

Although CD methods theoretically offer superior representational power by modeling both temporal and variable dependencies, their lack of prior knowledge about variable dependencies makes them prone to overfitting, especially with limited data~\cite{scalinglaw}. They may struggle to distinguish meaningful relationships from spurious correlations, leading to suboptimal generalization. This raises a key question: can we develop an explicit variable dependency modeling mechanism to enhance both many-to-one and many-to-many forecasting? To address this, we propose CrossLinear, which integrates cross-correlation embedding to incorporate time-invariant variable dependencies into the endogenous variable. Its single-layer design ensures effective modeling while mitigating overfitting by representing only direct variable dependencies.

\subsection{Forecasting with Exogenous Variables}
Although exogenous variable integration in time series forecasting has been underexplored in deep learning, it is a well-established statistical paradigm. ARIMA variants, such as ARIMAX and SARIMAX~\cite{arima}, incorporate regression terms to enhance forecasting. In deep learning, TiDE~\cite{tide} models both static and dynamic exogenous variables. However, its point-wise dependency modeling struggles to account for lead-lag effects~\cite{leadlag}. Recently, TimeXer~\cite{timexer} addresses this using self-attention and cross-attention, yet its variate-wise embedding limits variable dependency modeling and leads to poor performance with missing data.

The proposed CrossLinear method employs one-dimensional (1D) convolution to capture dependencies, adjusts for lead-lag effects, and ensures robustness with missing data by maintaining equal lengths for endogenous and embedded variables. Its cross-correlation embedding module can be integrated seamlessly into existing models, enhancing their forecasting performance.

\section{Methodology}
\subsection{Problem Formulation}
A univariate time series is defined as \(\mathbf{X}_{1:T,1} = \{x_1,\ x_2,\ \dots,\ x_T\} \in \mathbb{R}^{1 \times T}\), where \(T\) denotes the number of time points in the lookback window, and \(x_i \in \mathbb{R}\) represents the observation at the \(i^{\text{th}}\) time point. Similarly, a multivariate time series is defined as \(\mathbf{X}_{1:T,N} = \{\mathbf{x}_1,\ \mathbf{x}_2,\ \dots,\ \mathbf{x}_T\} \in \mathbb{R}^{N \times T}\), where \(N\) denotes the number of variables. Here, \(\mathbf{x}_i = \{v_{i,1},\ v_{i,2},\ \dots,\ v_{i,N}\}^\top \in \mathbb{R}^{N \times 1}\) represents the observations of all \(N\) variables at the \(i^{\text{th}}\) time point, and \(v_{i,j}\) denotes the observation value of the \(j^{\text{th}}\) variable at the \(i^{\text{th}}\) time point.

In the context of time series forecasting with exogenous variables, given an endogenous variable \(\mathbf{X}_{1:T,1}^{endo} \in \mathbb{R}^{1 \times T}\) and exogenous variables \(\mathbf{X}_{1:T,N-1}^{exo} \in \mathbb{R}^{(N-1) \times T}\), the objective of the forecaster \(\mathcal{M}\) is to predict the future \(S\)-step values of the endogenous variable, denoted as \(\hat{\mathbf{X}}_{T+1:T+S,1}^{endo} \in \mathbb{R}^{1 \times S}\), as shown below:

\begin{equation}
\hat{\mathbf{X}}_{T+1:T+S,1}^{endo} = \mathcal{M}(\mathbf{X}_{1:T,1}^{endo},\ \mathbf{X}_{1:T,N-1}^{exo})
\end{equation}

On the other hand, in the multivariate forecasting paradigm, the forecaster \(\mathcal{M}^{\prime}\) aims to predict the future \(S\)-step values for both the endogenous and exogenous variables:

\begin{equation}
\hat{\mathbf{X}}_{T+1:T+S,1}^{endo},\ \hat{\mathbf{X}}_{T+1:T+S,N-1}^{exo}= \mathcal{M}^{\prime}(\mathbf{X}_{1:T,1}^{endo},\ \mathbf{X}_{1:T,N-1}^{exo})
\end{equation}
where \(\hat{\mathbf{X}}_{T+1:T+S,N-1}^{exo}\in\mathbb{R}^{(N-1)\times S}\) represents the predicted values of the exogenous variables.

\subsection{Structure Overview}
As illustrated in Figure \ref{fig: Model Architecture}, the backbone of CrossLinear consists of the following key steps: (1) \textbf{Cross-Correlation Embedding}: a plug-and-play module designed to capture dependencies among variables, (2) \textbf{Patch Embedding}: a mechanism for extracting short-term temporal dependencies, and (3) \textbf{Forecasting Head}: a linear component that models long-term temporal dependencies.

In addition, to enhance training stability and mitigate the non-stationarity~\cite{nonstationarity_1, nonstationarity_2}, two modules, \textbf{Instance Normalization} and \textbf{De-Normalization} (from RevIN~\cite{revin}) are incorporated as preprocessing and postprocessing steps.

\subsection{Normalization and De-Normalization}
To address the issue of non-stationarity, a parameter-free RevIN module is employed, ensuring that the inputs to the backbone network maintain a stable distribution. This module has been widely adopted in various models, such as TimeXer, iTransformer and RLinear. The normalized endogenous and exogenous variables (\textit{i.e.}, \(\mathbf{X}_{1:T,1}^{endo*}\) and \(\mathbf{X}_{1:T,N-1}^{exo*}\)), the output of the backbone model \(\hat{\mathbf{X}}_{T+1:T+S,1}^{endo*}\), and the final output \(\hat{\mathbf{X}}_{T+1:T+S,1}^{endo}\) are defined as follows:

\begin{align}
\mathbf{X}_{1:T,1}^{endo*},\ \mu,\ \sigma &= \mathrm{InstanceNorm}(\mathbf{X}_{1:T,1}^{endo}) \label{2}\\
\mathbf{X}_{1:T,N-1}^{exo*} &= \mathrm{InstanceNorm}(\mathbf{X}_{1:T,N-1}^{exo}) \label{3}\\
\hat{\mathbf{X}}_{T+1:T+S,1}^{endo*} &= \mathcal{M}_{\mathrm{backbone}}(\mathbf{X}_{1:T,1}^{endo*},\ \mathbf{X}_{1:T,N-1}^{exo*}) \label{4}\\
\hat{\mathbf{X}}_{T+1:T+S,1}^{endo} &= \mathrm{De\mbox{-}Norm}(\hat{\mathbf{X}}_{T+1:T+S,1}^{endo*},\ \mu,\ \sigma)
\end{align}
where \(\mu\) and \(\sigma\) denote the mean and variance of the input endogenous variable, respectively, while \(\mathrm{InstanceNorm(\cdot)}\) and \(\mathrm{De\mbox{-}Norm(\cdot)}\) correspond to instance normalization and de-normalization. Besides, \(\mathcal{M}_{\mathrm{backbone}}(\cdot)\) represents the backbone network, excluding the normalization and de-normalization components.

\subsection{Cross-Correlation Embedding}
Existing models that explicitly model variable dependencies, such as Transformer-based models and graph neural networks, often suffer from overfitting~\cite{overfitting_1}. This overfitting primarily stems from the complexity and temporal dynamics of inter-variable dependencies, compounded by the limitations of insufficient or sparse data sources~\cite{msgnet,overfitting_2,univaraite_multivariate_3}. Moreover, the intricate dependency modeling employed by these models further exacerbates the risk of overfitting. To address these challenges, CrossLinear adopts a single-layer 1D convolution to efficiently extract variable dependencies while maintaining model simplicity. The result, \(\mathbf{X}_{1:T,1}^{cross}\), encodes embedded cross-correlation variables that capture essential inter-variable dependencies. To further enhance stability and mitigate overfitting, a residual structure is introduced, integrating both the original endogenous variable and the embedding. This fusion forms the final embedding vector, \(\mathbf{X}_{1:T,1}^{emb}\), which effectively preserves critical variable dependencies while reducing the risk of overfitting.

\begin{align}
\mathbf{X}_{1:T,1}^{cross} &= \mathrm{Conv1D}(\mathrm{Stack}(\mathbf{X}_{1:T,N-1}^{exo*},\ \mathbf{X}_{1:T,1}^{endo*})) \label{6}\\
\mathbf{X}_{1:T,1}^{emb} &= \alpha \cdot \mathbf{X}_{1:T,1}^{endo*} + (1-\alpha) \cdot \mathbf{X}_{1:T,1}^{cross}
\end{align}
where \(\mathrm{Stack(\cdot)}\) concatenates the endogenous and exogenous variables along the time dimension, forming an \(N \times T\) matrix. \(\mathrm{Conv1D(\cdot)}\) applies a 1D convolution to extract dependencies, while \(\alpha\) is a learnable parameter that balances the contributions of the endogenous variable and the cross-correlation variables.

\textbf{Note}: In theory, \(\mathbf{X}_{1:T,1}^{emb}\) and \(\mathbf{X}_{1:T,1}^{cross}\) are mathematically equivalent (see Appendix \ref{sec: ablation study appendix} for proof). While, treating endogenous and cross-correlation embeddings equally often prevents the model from learning appropriate representations, as discussed in Section \ref{sec: ablation study}.

\subsection{Patch Embedding, Positional Embedding, and Forecasting Head}
Patch embedding, originally from Vision Transformer~\cite{vit}, is widely employed in both Transformer- and Linear-based models, such as PatchTST~\cite{patchtst} and PatchMLP~\cite{patchmlp}, to capture short-term temporal dependencies, reduce the number of parameters, and mitigate overfitting~\cite{timexer}. The patching process is defined as follows:

\begin{equation}
\mathbf{p}_1^{endo},\ \mathbf{p}_2^{endo},\ \dots,\ \mathbf{p}_k^{endo} = \mathrm{Patchify}(\mathbf{X}_{1:T,1}^{emb})
\end{equation}
where \(p\) represents the patch length, \(k = \lceil T / p \rceil\) denotes the total number of patches, and each patch \(\mathbf{p}_i^{endo} \in \mathbb{R}^{1 \times p}\) corresponds to a segment of the input sequence.

To provide positional information and enhance robustness, positional embedding, which is commonly used in Transformer architectures, is incorporated. The forecasting head is then applied for capturing long-term temporal dependencies and generating the backbone model’s output, \(\hat{\mathbf{X}}_{T+1:T+S,1}^{endo*}\):

\begin{align}
\mathbf{P}^{endo} &= \beta \cdot \mathrm{Projection_1}(\mathbf{p}_1^{endo},\ \dots,\ \mathbf{p}_k^{endo}) + (1-\beta) \cdot \mathbf{PE} \label{9}\\
\hat{\mathbf{X}}_{T+1:T+S,1}^{endo*} &= \mathrm{Projection_2}(\mathrm{Concat}(\mathbf{P}^{endo}))
\end{align}
where the patches are first mapped into a \(d\)-dimensional space (\(d\) is a hyperparameter) through \(\mathrm{Projection_1}(\cdot)\). These embeddings are then summed with the positional embedding \(\mathbf{PE} \in \mathbb{R}^{k \times d}\), weighted by the learnable parameter \(\beta\). Finally, these embeddings are concatenated and then processed by \(\mathrm{Projection_2}(\cdot)\) to generate the final output.

\begin{table*}[!htbp]
\caption{Statistics of datasets. ``Exo num'' means the number of exogenous variables. The dataset size is organized in (Train, Validation, Test).}
\label{table: dataset}
\resizebox{1\linewidth}{!}{
\begin{tabular}{c|c|c|c|c|l}
\toprule
Dataset & Domain & Exo Num & Frequency & Dataset Size & Description \\ \midrule
Electricity & Electricity & 320 & 1 Hour & (18317, 2633, 5261) & Electricity records the electricity consumption in kWh every 1   hour from 2012 to 2014. \\
Weather & Weather & 20 & 10 Minutes & (36792, 5271, 10540) & Recorded every for the whole year 2020, which contains 21   meteorological indicators. \\
ETTh & Electricity & 6 & 1 Hour & (8545, 2881, 2881) & Power transformer, comprising seven indicators such as oil   temperature and useful load. \\
ETTm & Electricity & 6 & 15 Minutes & (34465, 11521, 11521) & Power transformer, comprising seven indicators such as oil   temperature and useful load. \\
Traffic & Traffic & 861 & 1 Hour & (12185, 1757, 3509) & Road occupancy rates measured by 862 sensors on San Francisco   Bay area freeways. \\
NP & Electricity & 2 & 1 Hour & (36500, 5219, 10460) & European power market of the Nordic countries, 2013 to 2018. \\
PJM & Electricity & 2 & 1 Hour & (36500, 5219, 10460) & American power market of Pennsylvania–New Jersey–Maryland (PJM), from 2013 to 2018. \\
BE & Electricity & 2 & 1 Hour & (36500, 5219, 10460) & Day-ahead electricity market in Belgium, from 2011 to 2016. \\
FR & Electricity & 2 & 1 Hour & (36500, 5219, 10460) & Day-ahead electricity market in France, from 2011 to 2016. \\
DE & Electricity & 2 & 1 Hour & (36500, 5219, 10460) & Day-ahead electricity market in Gemany, from 2012 to 2017. \\ \bottomrule
\end{tabular}}
\end{table*}

\subsection{Extension to Multivariate Forecasting}
For multivariate time series forecasting, the task can be interpreted as the aggregation of \(N\) individual many-to-one forecasting tasks. While parallel processing is possible, weight-sharing offers a more resource-efficient solution during both training and inference. To accommodate this approach, the number of output channels in the cross-correlation embedding module is adjusted, enabling the generation of distinct cross-correlation variables for each endogenous variable. Given the computational efficiency of 1D convolution, this adjustment introduces only minimal overhead. Subsequently, weight-sharing is applied across the patch embedding, positional embedding, and forecasting head, facilitating efficient and scalable multivariate forecasting.

\subsection{Forecasting Loss}
During model training, we employ the L2 loss as the objective function. For forecasting with exogenous variables, the goal is to minimize the discrepancy between the forecasting values of endogenous variable, \(\hat{\mathbf{X}}_{T+1:T+S,1}^{endo}\), and the ground truth, \(\mathbf{X}_{T+1:T+S,1}^{endo}\). The loss function \(\mathcal{L}\) is defined as:

\begin{equation}
\mathcal{L} = \Vert\hat{\mathbf{X}}_{T+1:T+S,1}^{endo} - \mathbf{X}_{T+1:T+S,1}^{endo}\Vert_{2}^{2}
\end{equation}

For multivariate time series forecasting tasks, it is crucial to ensure the precision of both endogenous and exogenous variables. Therefore, the loss function \(\mathcal{L}\) is extended as follows:

\begin{equation}
\mathcal{L} = \Vert\hat{\mathbf{X}}_{T+1:T+S,1}^{endo} - \mathbf{X}_{T+1:T+S,1}^{endo}\Vert_{2}^{2} + \Vert\hat{\mathbf{X}}_{T+1:T+S,N-1}^{exo} - \mathbf{X}_{T+1:T+S,N-1}^{exo}\Vert_{2}^{2}
\end{equation}

\section{Experiments and Results}
In our study, we utilized a comprehensive collection of 12 datasets, following experimental configurations consistent with TimeXer~\cite{timexer}. We evaluated 10 state-of-the-art models, encompassing a diverse range of architectures, including Transformer-based, Linear-based, CNN-based and GNN-based approaches. These methodological choices were carefully made to ensure the robustness and reliability of our experimental results.

\textbf{Datasets.}
To evaluate the effectiveness of long-term forecasting, we conducted a series of experiments using 7 real-world datasets~\cite{long_dataset}, including \textbf{ECL}, \textbf{Weather}, \textbf{ETTh1}, \textbf{ETTh2}, \textbf{ETTm1}, \textbf{ETTm2}, and \textbf{Traffic}. For short-term forecasting, we utilized the \textbf{EPF} dataset~\cite{short_dataset}, which comprises 5 subsets specifically designed for short-term forecasting. These datasets span multiple domains, such as weather, electricity prices, and traffic, representing key data types commonly analyzed in modern time series research, as shown in Table \ref{table: dataset}. This broad coverage enhances the generalizability and practical relevance of our findings across various applications.

\textbf{Baselines.}
We include 10 state-of-the-art deep forecasting models, including (1) \textbf{Transformer-based}: TimeXer~\cite{timexer}, iTransformer~\cite{itransformer}, PatchTST~\cite{patchtst}, Autoformer~\cite{autoformer}, (2) \textbf{Linear-based}: SparseTSF~\cite{sparsetsf}, RLinear~\cite{rlinear}, DLinear~\cite{dlinear}, TiDE~\cite{tide}, (3) \textbf{CNN-based}: TimesNet~\cite{timesnet}, and (4) \textbf{GNN-based}: MSGNet~\cite{msgnet}. It is worth noting that TimeXer and TiDE are recently developed advanced predictors specifically designed for exogenous variables.

\textbf{Setup.}
Our experimental setup follows TimeXer~\cite{timexer}. For long-term forecasting, we fix the lookback window \(T\) at 96 and set the forecasting horizons \(S\) to \{96, 192, 336, 720\}, which allows us to comprehensively evaluate the model across various temporal scales. For short-term forecasting, we use a lookback window \(T\) of 168 and a forecasting horizon \(S\) of 24, in line with standard practices for short-term prediction tasks~\cite{nbeatsx}. The model's performance is evaluated using two primary metrics: Mean Squared Error (MSE) and Mean Absolute Error (MAE), ensuring a robust assessment of both overall precision and the magnitude of deviations.

Further details of the experimental setup are provided in Appendix~\ref{sec: details of experiments}.

\subsection{Main Results}
Using the above mentioned setup, we next present our findings.

\begin{table*}[!htbp]
\caption{Full results of the forecasting task with exogenous variables. ``Avg.'' refers to the average results across all five datasets. The best results are highlighted in bold. Note: The Weather dataset is compared based on four decimal places.}
\label{table: ms forecast}
\resizebox{1\linewidth}{!}{
\begin{tabular}{cc|cc|cc|cc|cc|cc|cc|cc|cc|cc|cc|cc}
\toprule
\multicolumn{2}{c|}{\multirow{2}{*}{Model}} & \multicolumn{2}{c|}{\textbf{CrossLinear}} & \multicolumn{2}{c|}{TimeXer} & \multicolumn{2}{c|}{iTransformer} & \multicolumn{2}{c|}{MSGNet} & \multicolumn{2}{c|}{SparseTSF} & \multicolumn{2}{c|}{RLinear} & \multicolumn{2}{c|}{PatchTST} & \multicolumn{2}{c|}{TiDE} & \multicolumn{2}{c|}{TimesNet} & \multicolumn{2}{c|}{DLinear} & \multicolumn{2}{c}{Autoformer} \\
\multicolumn{2}{c|}{} & \multicolumn{2}{c|}{\textbf{(Ours)}} & \multicolumn{2}{c|}{(2024)} & \multicolumn{2}{c|}{(2024)} & \multicolumn{2}{c|}{(2024)} & \multicolumn{2}{c|}{(2024)} & \multicolumn{2}{c|}{(2023)} & \multicolumn{2}{c|}{(2023)} & \multicolumn{2}{c|}{(2023)} & \multicolumn{2}{c|}{(2023)} & \multicolumn{2}{c|}{(2023)} & \multicolumn{2}{c}{(2021)} \\
\multicolumn{2}{c|}{Metric} & MSE & MAE & MSE & MAE & MSE & MAE & MSE & MAE & MSE & MAE & MSE & MAE & MSE & MAE & MSE & MAE & MSE & MAE & MSE & MAE & MSE & MAE \\ \midrule
\multicolumn{1}{c|}{\multirow{5}{*}{ECL}} & 96 & \textbf{0.251} & \textbf{0.359} & 0.261 & 0.366 & 0.299 & 0.403 & 0.349 & 0.442 & 0.332 & 0.404 & 0.433 & 0.480 & 0.339 & 0.412 & 0.405 & 0.459 & 0.342 & 0.437 & 0.387 & 0.451 & 0.432 & 0.502 \\
\multicolumn{1}{c|}{} & 192 & \textbf{0.294} & \textbf{0.381} & 0.316 & 0.397 & 0.321 & 0.413 & 0.400 & 0.477 & 0.335 & 0.402 & 0.407 & 0.461 & 0.361 & 0.425 & 0.383 & 0.442 & 0.384 & 0.461 & 0.365 & 0.436 & 0.492 & 0.492 \\
\multicolumn{1}{c|}{} & 336 & \textbf{0.343} & \textbf{0.416} & 0.367 & 0.429 & 0.379 & 0.446 & 0.399 & 0.476 & 0.378 & 0.432 & 0.440 & 0.481 & 0.393 & 0.440 & 0.418 & 0.464 & 0.439 & 0.493 & 0.391 & 0.453 & 0.508 & 0.548 \\
\multicolumn{1}{c|}{} & 720 & 0.403 & 0.465 & \textbf{0.365} & \textbf{0.439} & 0.461 & 0.504 & 0.400 & 0.476 & 0.444 & 0.486 & 0.495 & 0.523 & 0.482 & 0.507 & 0.471 & 0.507 & 0.473 & 0.514 & 0.428 & 0.487 & 0.547 & 0.569 \\
\multicolumn{1}{c|}{} & Avg. & \textbf{0.323} & \textbf{0.405} & 0.327 & 0.408 & 0.365 & 0.442 & 0.387 & 0.468 & 0.372 & 0.431 & 0.444 & 0.486 & 0.394 & 0.446 & 0.419 & 0.468 & 0.410 & 0.476 & 0.393 & 0.457 & 0.495 & 0.528 \\ \midrule
\multicolumn{1}{c|}{\multirow{5}{*}{Weather}} & 96 & 0.001 & 0.028 & 0.001 & 0.027 & 0.001 & 0.026 & 0.001 & 0.028 & 0.002 & 0.030 & \textbf{0.001} & \textbf{0.025} & 0.001 & 0.027 & \textbf{0.001} & \textbf{0.025} & 0.002 & 0.029 & 0.006 & 0.062 & 0.007 & 0.066 \\
\multicolumn{1}{c|}{} & 192 & 0.002 & 0.030 & 0.002 & 0.030 & 0.002 & 0.029 & 0.002 & 0.031 & 0.002 & 0.037 & \textbf{0.001} & \textbf{0.028} & 0.002 & 0.030 & \textbf{0.001} & \textbf{0.028} & 0.002 & 0.031 & 0.006 & 0.066 & 0.007 & 0.061 \\
\multicolumn{1}{c|}{} & 336 & 0.002 & 0.032 & 0.002 & 0.031 & 0.002 & 0.031 & 0.002 & 0.031 & 0.003 & 0.044 & \textbf{0.002} & \textbf{0.029} & 0.002 & 0.032 & \textbf{0.002} & \textbf{0.029} & 0.002 & 0.031 & 0.006 & 0.068 & 0.007 & 0.062 \\
\multicolumn{1}{c|}{} & 720 & 0.002 & 0.036 & 0.002 & 0.036 & 0.002 & 0.036 & 0.002 & 0.035 & 0.002 & 0.034 & \textbf{0.002} & \textbf{0.033} & 0.002 & 0.036 & \textbf{0.002} & \textbf{0.033} & 0.381 & 0.368 & 0.007 & 0.070 & 0.005 & 0.053 \\
\multicolumn{1}{c|}{} & Avg. & 0.002 & 0.031 & 0.002 & 0.031 & 0.002 & 0.031 & 0.002 & 0.031 & 0.002 & 0.036 & \textbf{0.002} & \textbf{0.029} & 0.002 & 0.031 & \textbf{0.002} & \textbf{0.029} & 0.097 & 0.115 & 0.006 & 0.066 & 0.006 & 0.060 \\ \midrule
\multicolumn{1}{c|}{\multirow{5}{*}{ETTh1}} & 96 & \textbf{0.055} & \textbf{0.178} & 0.057 & 0.181 & 0.057 & 0.183 & 0.058 & 0.183 & 0.063 & 0.199 & 0.059 & 0.185 & \textbf{0.055} & \textbf{0.178} & 0.059 & 0.184 & 0.059 & 0.188 & 0.065 & 0.188 & 0.119 & 0.263 \\
\multicolumn{1}{c|}{} & 192 & 0.072 & 0.205 & \textbf{0.071} & \textbf{0.204} & 0.074 & 0.209 & 0.076 & 0.212 & 0.080 & 0.224 & 0.078 & 0.214 & 0.072 & 0.206 & 0.078 & 0.214 & 0.080 & 0.217 & 0.088 & 0.222 & 0.132 & 0.286 \\
\multicolumn{1}{c|}{} & 336 & 0.082 & 0.226 & \textbf{0.080} & 0.223 & 0.084 & 0.223 & 0.081 & \textbf{0.221} & 0.091 & 0.241 & 0.093 & 0.240 & 0.087 & 0.231 & 0.093 & 0.240 & 0.083 & 0.224 & 0.110 & 0.257 & 0.126 & 0.278 \\
\multicolumn{1}{c|}{} & 720 & \textbf{0.080} & \textbf{0.225} & 0.084 & 0.229 & 0.084 & 0.229 & 0.089 & 0.236 & 0.081 & 0.229 & 0.106 & 0.256 & 0.098 & 0.247 & 0.104 & 0.255 & 0.083 & 0.231 & 0.202 & 0.371 & 0.143 & 0.299 \\
\multicolumn{1}{c|}{} & Avg. & \textbf{0.072} & \textbf{0.208} & 0.073 & 0.209 & 0.075 & 0.211 & 0.076 & 0.213 & 0.079 & 0.224 & 0.084 & 0.224 & 0.078 & 0.215 & 0.083 & 0.223 & 0.076 & 0.215 & 0.116 & 0.259 & 0.130 & 0.282 \\ \midrule
\multicolumn{1}{c|}{\multirow{5}{*}{ETTh2}} & 96 & \textbf{0.131} & \textbf{0.279} & 0.132 & 0.280 & 0.137 & 0.287 & 0.144 & 0.295 & 0.137 & 0.289 & 0.136 & 0.286 & 0.136 & 0.285 & 0.136 & 0.285 & 0.159 & 0.310 & 0.135 & 0.282 & 0.184 & 0.335 \\
\multicolumn{1}{c|}{} & 192 & \textbf{0.178} & \textbf{0.332} & 0.181 & 0.333 & 0.187 & 0.341 & 0.192 & 0.346 & 0.183 & 0.339 & 0.187 & 0.339 & 0.185 & 0.337 & 0.187 & 0.339 & 0.196 & 0.351 & 0.188 & 0.335 & 0.214 & 0.364 \\
\multicolumn{1}{c|}{} & 336 & \textbf{0.214} & \textbf{0.370} & 0.223 & 0.377 & 0.221 & 0.376 & 0.217 & 0.373 & 0.222 & 0.379 & 0.231 & 0.384 & 0.217 & 0.373 & 0.231 & 0.384 & 0.232 & 0.385 & 0.238 & 0.385 & 0.269 & 0.405 \\
\multicolumn{1}{c|}{} & 720 & \textbf{0.220} & \textbf{0.376} & \textbf{0.220} & \textbf{0.376} & 0.253 & 0.403 & 0.229 & 0.385 & 0.236 & 0.392 & 0.267 & 0.417 & 0.229 & 0.384 & 0.267 & 0.417 & 0.254 & 0.403 & 0.336 & 0.475 & 0.303 & 0.440 \\
\multicolumn{1}{c|}{} & Avg. & \textbf{0.186} & \textbf{0.339} & 0.189 & 0.342 & 0.199 & 0.352 & 0.195 & 0.350 & 0.194 & 0.350 & 0.205 & 0.356 & 0.192 & 0.345 & 0.205 & 0.356 & 0.210 & 0.362 & 0.224 & 0.369 & 0.242 & 0.386 \\ \midrule
\multicolumn{1}{c|}{\multirow{5}{*}{ETTm1}} & 96 & \textbf{0.028} & \textbf{0.125} & \textbf{0.028} & \textbf{0.125} & 0.029 & 0.128 & \textbf{0.028} & 0.126 & 0.030 & 0.129 & 0.030 & 0.129 & 0.029 & 0.126 & 0.030 & 0.129 & 0.029 & 0.128 & 0.034 & 0.135 & 0.097 & 0.251 \\
\multicolumn{1}{c|}{} & 192 & 0.044 & \textbf{0.158} & \textbf{0.043} & \textbf{0.158} & 0.045 & 0.163 & 0.044 & 0.160 & 0.045 & 0.161 & 0.044 & 0.160 & 0.045 & 0.160 & 0.044 & 0.160 & 0.044 & 0.160 & 0.055 & 0.173 & 0.062 & 0.197 \\
\multicolumn{1}{c|}{} & 336 & \textbf{0.057} & \textbf{0.183} & 0.058 & 0.185 & 0.060 & 0.190 & 0.059 & 0.187 & 0.058 & 0.185 & \textbf{0.057} & 0.184 & 0.058 & 0.184 & \textbf{0.057} & 0.184 & 0.061 & 0.190 & 0.078 & 0.210 & 0.083 & 0.230 \\
\multicolumn{1}{c|}{} & 720 & 0.080 & \textbf{0.215} & \textbf{0.079} & 0.217 & \textbf{0.079} & 0.218 & 0.083 & 0.220 & 0.081 & 0.221 & 0.080 & 0.217 & 0.082 & 0.221 & 0.080 & 0.217 & 0.083 & 0.223 & 0.098 & 0.234 & 0.100 & 0.245 \\
\multicolumn{1}{c|}{} & Avg. & \textbf{0.052} & \textbf{0.170} & \textbf{0.052} & 0.171 & 0.053 & 0.175 & 0.054 & 0.173 & 0.053 & 0.174 & 0.053 & 0.173 & 0.053 & 0.173 & 0.053 & 0.173 & 0.054 & 0.175 & 0.066 & 0.188 & 0.085 & 0.230 \\ \midrule
\multicolumn{1}{c|}{\multirow{5}{*}{ETTm2}} & 96 & \textbf{0.064} & \textbf{0.180} & 0.067 & 0.188 & 0.071 & 0.194 & 0.074 & 0.203 & 0.068 & 0.190 & 0.074 & 0.199 & 0.068 & 0.188 & 0.073 & 0.199 & 0.073 & 0.200 & 0.072 & 0.195 & 0.133 & 0.282 \\
\multicolumn{1}{c|}{} & 192 & \textbf{0.098} & \textbf{0.232} & 0.101 & 0.236 & 0.108 & 0.247 & 0.108 & 0.250 & 0.103 & 0.238 & 0.104 & 0.241 & 0.100 & 0.236 & 0.104 & 0.241 & 0.106 & 0.247 & 0.105 & 0.240 & 0.143 & 0.294 \\
\multicolumn{1}{c|}{} & 336 & \textbf{0.128} & \textbf{0.271} & 0.130 & 0.275 & 0.140 & 0.288 & 0.141 & 0.288 & 0.133 & 0.277 & 0.131 & 0.276 & \textbf{0.128} & \textbf{0.271} & 0.131 & 0.276 & 0.150 & 0.296 & 0.136 & 0.280 & 0.156 & 0.308 \\
\multicolumn{1}{c|}{} & 720 & 0.183 & 0.332 & 0.182 & 0.332 & 0.188 & 0.340 & 0.188 & 0.339 & 0.185 & 0.334 & \textbf{0.180} & \textbf{0.329} & 0.185 & 0.335 & \textbf{0.180} & \textbf{0.329} & 0.186 & 0.338 & 0.191 & 0.335 & 0.184 & 0.333 \\
\multicolumn{1}{c|}{} & Avg. & \textbf{0.118} & \textbf{0.254} & 0.120 & 0.258 & 0.127 & 0.267 & 0.128 & 0.270 & 0.122 & 0.260 & 0.122 & 0.261 & 0.120 & 0.258 & 0.122 & 0.261 & 0.129 & 0.271 & 0.126 & 0.263 & 0.154 & 0.305 \\ \midrule
\multicolumn{1}{c|}{\multirow{5}{*}{Traffic}} & 96 & \textbf{0.149} & \textbf{0.223} & 0.151 & 0.224 & 0.156 & 0.236 & 0.189 & 0.284 & 0.204 & 0.288 & 0.350 & 0.431 & 0.176 & 0.253 & 0.350 & 0.430 & 0.154 & 0.249 & 0.268 & 0.351 & 0.250 & 0.343 \\
\multicolumn{1}{c|}{} & 192 & \textbf{0.149} & \textbf{0.225} & 0.152 & 0.229 & 0.156 & 0.237 & 0.204 & 0.300 & 0.187 & 0.267 & 0.314 & 0.404 & 0.162 & 0.243 & 0.230 & 0.315 & 0.164 & 0.255 & 0.302 & 0.387 & 0.294 & 0.396 \\
\multicolumn{1}{c|}{} & 336 & \textbf{0.148} & \textbf{0.229} & 0.150 & 0.232 & 0.154 & 0.243 & 0.215 & 0.318 & 0.186 & 0.264 & 0.305 & 0.399 & 0.164 & 0.248 & 0.220 & 0.208 & 0.167 & 0.259 & 0.298 & 0.384 & 0.322 & 0.416 \\
\multicolumn{1}{c|}{} & 720 & \textbf{0.161} & \textbf{0.247} & 0.172 & 0.253 & 0.177 & 0.268 & 0.245 & 0.348 & 0.197 & 0.279 & 0.328 & 0.415 & 0.189 & 0.267 & 0.243 & 0.329 & 0.197 & 0.292 & 0.340 & 0.416 & 0.307 & 0.414 \\
\multicolumn{1}{c|}{} & Avg. & \textbf{0.152} & \textbf{0.231} & 0.156 & 0.234 & 0.161 & 0.246 & 0.213 & 0.313 & 0.194 & 0.274 & 0.324 & 0.412 & 0.173 & 0.253 & 0.240 & 0.326 & 0.171 & 0.264 & 0.323 & 0.404 & 0.293 & 0.392 \\ \midrule
\multicolumn{1}{c|}{\multirow{6}{*}{EPF}} & NP & \textbf{0.232} & \textbf{0.268} & 0.236 & \textbf{0.268} & 0.265 & 0.300 & 0.288 & 0.306 & 0.310 & 0.323 & 0.335 & 0.340 & 0.267 & 0.284 & 0.335 & 0.340 & 0.250 & 0.289 & 0.309 & 0.321 & 0.402 & 0.398 \\
\multicolumn{1}{c|}{} & PJM & \textbf{0.092} & 0.193 & 0.093 & \textbf{0.192} & 0.097 & 0.197 & 0.118 & 0.226 & 0.115 & 0.229 & 0.124 & 0.229 & 0.106 & 0.209 & 0.124 & 0.228 & 0.097 & 0.195 & 0.108 & 0.215 & 0.168 & 0.267 \\
\multicolumn{1}{c|}{} & BE & \textbf{0.372} & 0.248 & 0.379 & \textbf{0.243} & 0.394 & 0.270 & 0.418 & 0.276 & 0.432 & 0.289 & 0.520 & 0.337 & 0.400 & 0.262 & 0.523 & 0.336 & 0.419 & 0.288 & 0.463 & 0.313 & 0.500 & 0.333 \\
\multicolumn{1}{c|}{} & FR & \textbf{0.384} & \textbf{0.207} & 0.385 & 0.208 & 0.439 & 0.233 & 0.424 & 0.238 & \textbf{0.384} & 0.223 & 0.507 & 0.290 & 0.411 & 0.220 & 0.510 & 0.290 & 0.431 & 0.234 & 0.429 & 0.260 & 0.519 & 0.295 \\
\multicolumn{1}{c|}{} & DE & \textbf{0.437} & \textbf{0.413} & 0.440 & 0.415 & 0.479 & 0.443 & 0.500 & 0.448 & 0.513 & 0.464 & 0.574 & 0.498 & 0.461 & 0.432 & 0.568 & 0.496 & 0.502 & 0.446 & 0.520 & 0.463 & 0.674 & 0.544 \\
\multicolumn{1}{c|}{} & Avg. & \textbf{0.303} & 0.266 & 0.307 & \textbf{0.265} & 0.335 & 0.289 & 0.349 & 0.299 & 0.351 & 0.305 & 0.412 & 0.339 & 0.330 & 0.282 & 0.412 & 0.338 & 0.340 & 0.290 & 0.366 & 0.314 & 0.453 & 0.368 \\ \midrule
\multicolumn{2}{c|}{1\(^{\mathrm{st}}\) Count} & 30 & 29 & 8 & 9 & 1 & 0 & 1 & 1 & 1 & 0 & 7 & 6 & 2 & 2 & 7 & 6 & 0 & 0 & 0 & 0 & 0 & 0 \\ \bottomrule
\end{tabular}}
\end{table*}

\begin{table*}[!htbp]
\caption{Full results of the multivariate forecasting task. ``Avg.'' refers to the average results across all datasets. The best results are highlighted in bold.}
\label{table: mm forecast}
\resizebox{1\linewidth}{!}{
\begin{tabular}{cc|cc|cc|cc|cc|cc|cc|cc|cc|cc|cc|cc}
\toprule
\multicolumn{2}{c|}{\multirow{2}{*}{Model}} & \multicolumn{2}{c|}{\textbf{CrossLinear}} & \multicolumn{2}{c|}{TimeXer} & \multicolumn{2}{c|}{iTransformer} & \multicolumn{2}{c|}{MSGNet} & \multicolumn{2}{c|}{SparseTSF} & \multicolumn{2}{c|}{RLinear} & \multicolumn{2}{c|}{PatchTST} & \multicolumn{2}{c|}{TiDE} & \multicolumn{2}{c|}{TimesNet} & \multicolumn{2}{c|}{DLinear} & \multicolumn{2}{c}{Autoformer} \\
\multicolumn{2}{c|}{} & \multicolumn{2}{c|}{\textbf{(Ours)}} & \multicolumn{2}{c|}{(2024)} & \multicolumn{2}{c|}{(2024)} & \multicolumn{2}{c|}{(2024)} & \multicolumn{2}{c|}{(2024)} & \multicolumn{2}{c|}{(2023)} & \multicolumn{2}{c|}{(2023)} & \multicolumn{2}{c|}{(2023)} & \multicolumn{2}{c|}{(2023)} & \multicolumn{2}{c|}{(2023)} & \multicolumn{2}{c}{(2021)} \\
\multicolumn{2}{c|}{Metric} & MSE & MAE & MSE & MAE & MSE & MAE & MSE & MAE & MSE & MAE & MSE & MAE & MSE & MAE & MSE & MAE & MSE & MAE & MSE & MAE & MSE & MAE \\ \midrule
\multicolumn{1}{c|}{\multirow{5}{*}{ECL}} & 96 & \textbf{0.139} & \textbf{0.237} & 0.140 & 0.242 & 0.148 & 0.240 & 0.165 & 0.274 & 0.197 & 0.269 & 0.201 & 0.281 & 0.195 & 0.285 & 0.237 & 0.329 & 0.168 & 0.272 & 0.197 & 0.282 & 0.201 & 0.317 \\
\multicolumn{1}{c|}{} & 192 & \textbf{0.157} & 0.254 & \textbf{0.157} & 0.256 & 0.162 & \textbf{0.253} & 0.184 & 0.292 & 0.195 & 0.272 & 0.201 & 0.283 & 0.199 & 0.289 & 0.236 & 0.330 & 0.184 & 0.289 & 0.196 & 0.285 & 0.222 & 0.334 \\
\multicolumn{1}{c|}{} & 336 & \textbf{0.176} & 0.275 & \textbf{0.176} & 0.275 & 0.178 & \textbf{0.269} & 0.195 & 0.302 & 0.209 & 0.287 & 0.215 & 0.298 & 0.215 & 0.305 & 0.249 & 0.344 & 0.198 & 0.300 & 0.209 & 0.301 & 0.231 & 0.338 \\
\multicolumn{1}{c|}{} & 720 & 0.221 & 0.315 & \textbf{0.211} & \textbf{0.306} & 0.225 & 0.317 & 0.231 & 0.332 & 0.251 & 0.321 & 0.257 & 0.331 & 0.256 & 0.337 & 0.284 & 0.373 & 0.220 & 0.320 & 0.245 & 0.333 & 0.254 & 0.361 \\
\multicolumn{1}{c|}{} & Avg. & 0.174 & \textbf{0.270} & \textbf{0.171} & \textbf{0.270} & 0.178 & \textbf{0.270} & 0.194 & 0.300 & 0.213 & 0.287 & 0.219 & 0.298 & 0.216 & 0.304 & 0.251 & 0.344 & 0.192 & 0.295 & 0.212 & 0.300 & 0.227 & 0.338 \\ \midrule
\multicolumn{1}{c|}{\multirow{5}{*}{Weather}} & 96 & \textbf{0.154} & \textbf{0.202} & 0.157 & 0.205 & 0.174 & 0.214 & 0.163 & 0.212 & 0.180 & 0.231 & 0.192 & 0.232 & 0.177 & 0.218 & 0.202 & 0.261 & 0.172 & 0.220 & 0.196 & 0.255 & 0.266 & 0.336 \\
\multicolumn{1}{c|}{} & 192 & \textbf{0.200} & \textbf{0.246} & 0.204 & 0.247 & 0.221 & 0.254 & 0.212 & 0.254 & 0.225 & 0.267 & 0.240 & 0.271 & 0.225 & 0.259 & 0.242 & 0.298 & 0.219 & 0.261 & 0.237 & 0.296 & 0.307 & 0.367 \\
\multicolumn{1}{c|}{} & 336 & \textbf{0.257} & \textbf{0.286} & 0.261 & 0.290 & 0.278 & 0.296 & 0.272 & 0.299 & 0.280 & 0.309 & 0.292 & 0.307 & 0.278 & 0.297 & 0.287 & 0.335 & 0.280 & 0.306 & 0.283 & 0.335 & 0.359 & 0.395 \\
\multicolumn{1}{c|}{} & 720 & \textbf{0.340} & 0.343 & \textbf{0.340} & \textbf{0.341} & 0.358 & 0.349 & 0.350 & 0.348 & 0.360 & 0.365 & 0.364 & 0.353 & 0.354 & 0.348 & 0.351 & 0.386 & 0.365 & 0.359 & 0.345 & 0.381 & 0.419 & 0.428 \\
\multicolumn{1}{c|}{} & Avg. & \textbf{0.238} & \textbf{0.269} & 0.241 & 0.271 & 0.258 & 0.279 & 0.249 & 0.278 & 0.261 & 0.293 & 0.272 & 0.291 & 0.259 & 0.281 & 0.271 & 0.320 & 0.259 & 0.287 & 0.265 & 0.317 & 0.338 & 0.382 \\ \midrule
\multicolumn{1}{c|}{\multirow{5}{*}{ETTh1}} & 96 & \textbf{0.374} & \textbf{0.393} & 0.382 & 0.403 & 0.386 & 0.405 & 0.390 & 0.411 & 0.376 & 0.395 & 0.386 & 0.395 & 0.414 & 0.419 & 0.479 & 0.464 & 0.384 & 0.402 & 0.386 & 0.400 & 0.449 & 0.459 \\
\multicolumn{1}{c|}{} & 192 & \textbf{0.422} & \textbf{0.424} & 0.429 & 0.435 & 0.441 & 0.436 & 0.442 & 0.442 & 0.426 & 0.429 & 0.437 & \textbf{0.424} & 0.460 & 0.445 & 0.525 & 0.492 & 0.436 & 0.429 & 0.437 & 0.432 & 0.500 & 0.482 \\
\multicolumn{1}{c|}{} & 336 & \textbf{0.459} & 0.447 & 0.468 & 0.448 & 0.487 & 0.458 & 0.480 & 0.468 & 0.470 & \textbf{0.441} & 0.479 & 0.446 & 0.501 & 0.466 & 0.565 & 0.515 & 0.491 & 0.469 & 0.481 & 0.459 & 0.521 & 0.496 \\
\multicolumn{1}{c|}{} & 720 & \textbf{0.467} & 0.465 & 0.469 & \textbf{0.461} & 0.503 & 0.491 & 0.494 & 0.488 & 0.483 & 0.469 & 0.481 & 0.470 & 0.500 & 0.488 & 0.594 & 0.558 & 0.521 & 0.500 & 0.519 & 0.516 & 0.514 & 0.512 \\
\multicolumn{1}{c|}{} & Avg. & \textbf{0.431} & \textbf{0.433} & 0.437 & 0.437 & 0.454 & 0.447 & 0.452 & 0.452 & 0.439 & \textbf{0.433} & 0.446 & 0.434 & 0.469 & 0.454 & 0.541 & 0.507 & 0.458 & 0.450 & 0.456 & 0.452 & 0.496 & 0.487 \\ \midrule
\multicolumn{1}{c|}{\multirow{5}{*}{ETTh2}} & 96 & \textbf{0.282} & \textbf{0.337} & 0.286 & 0.338 & 0.297 & 0.349 & 0.328 & 0.371 & 0.308 & 0.354 & 0.288 & 0.338 & 0.302 & 0.348 & 0.400 & 0.440 & 0.340 & 0.374 & 0.333 & 0.387 & 0.346 & 0.388 \\
\multicolumn{1}{c|}{} & 192 & \textbf{0.360} & \textbf{0.387} & 0.363 & 0.389 & 0.380 & 0.400 & 0.402 & 0.414 & 0.388 & 0.398 & 0.374 & 0.390 & 0.388 & 0.400 & 0.528 & 0.509 & 0.402 & 0.414 & 0.477 & 0.476 & 0.456 & 0.452 \\
\multicolumn{1}{c|}{} & 336 & \textbf{0.405} & \textbf{0.422} & 0.414 & 0.423 & 0.428 & 0.432 & 0.435 & 0.443 & 0.431 & 0.443 & 0.415 & 0.426 & 0.426 & 0.433 & 0.643 & 0.571 & 0.452 & 0.452 & 0.594 & 0.541 & 0.482 & 0.486 \\
\multicolumn{1}{c|}{} & 720 & 0.424 & 0.439 & \textbf{0.408} & \textbf{0.432} & 0.427 & 0.445 & 0.417 & 0.441 & 0.432 & 0.447 & 0.420 & 0.440 & 0.431 & 0.446 & 0.874 & 0.679 & 0.462 & 0.468 & 0.831 & 0.657 & 0.515 & 0.511 \\
\multicolumn{1}{c|}{} & Avg. & 0.368 & \textbf{0.396} & \textbf{0.367} & \textbf{0.396} & 0.383 & 0.407 & 0.396 & 0.417 & 0.390 & 0.411 & 0.374 & 0.398 & 0.387 & 0.407 & 0.611 & 0.550 & 0.414 & 0.427 & 0.559 & 0.515 & 0.450 & 0.459 \\ \midrule
\multicolumn{1}{c|}{\multirow{5}{*}{ETTm1}} & 96 & \textbf{0.311} & \textbf{0.354} & 0.318 & 0.356 & 0.334 & 0.368 & 0.319 & 0.366 & 0.349 & 0.376 & 0.355 & 0.376 & 0.329 & 0.367 & 0.364 & 0.387 & 0.338 & 0.375 & 0.345 & 0.372 & 0.505 & 0.475 \\
\multicolumn{1}{c|}{} & 192 & \textbf{0.352} & \textbf{0.379} & 0.362 & 0.383 & 0.387 & 0.391 & 0.376 & 0.397 & 0.380 & 0.393 & 0.391 & 0.392 & 0.367 & 0.385 & 0.398 & 0.404 & 0.374 & 0.387 & 0.380 & 0.389 & 0.553 & 0.496 \\
\multicolumn{1}{c|}{} & 336 & \textbf{0.381} & \textbf{0.401} & 0.395 & 0.407 & 0.426 & 0.420 & 0.417 & 0.422 & 0.410 & 0.413 & 0.424 & 0.415 & 0.399 & 0.410 & 0.428 & 0.425 & 0.410 & 0.411 & 0.413 & 0.413 & 0.621 & 0.537 \\
\multicolumn{1}{c|}{} & 720 & \textbf{0.439} & \textbf{0.439} & 0.452 & 0.441 & 0.491 & 0.459 & 0.481 & 0.458 & 0.473 & 0.447 & 0.487 & 0.450 & 0.454 & \textbf{0.439} & 0.487 & 0.461 & 0.478 & 0.450 & 0.474 & 0.453 & 0.671 & 0.561 \\
\multicolumn{1}{c|}{} & Avg. & \textbf{0.370} & \textbf{0.393} & 0.382 & 0.397 & 0.407 & 0.410 & 0.398 & 0.411 & 0.403 & 0.408 & 0.414 & 0.407 & 0.387 & 0.400 & 0.419 & 0.419 & 0.400 & 0.406 & 0.403 & 0.407 & 0.588 & 0.517 \\ \midrule
\multicolumn{1}{c|}{\multirow{5}{*}{ETTm2}} & 96 & \textbf{0.170} & \textbf{0.254} & 0.171 & 0.256 & 0.180 & 0.264 & 0.177 & 0.262 & 0.180 & 0.262 & 0.182 & 0.265 & 0.175 & 0.259 & 0.207 & 0.305 & 0.187 & 0.267 & 0.193 & 0.292 & 0.255 & 0.339 \\
\multicolumn{1}{c|}{} & 192 & \textbf{0.236} & \textbf{0.298} & 0.237 & 0.299 & 0.250 & 0.309 & 0.247 & 0.307 & 0.243 & 0.302 & 0.246 & 0.304 & 0.241 & 0.302 & 0.290 & 0.364 & 0.249 & 0.309 & 0.284 & 0.362 & 0.281 & 0.340 \\
\multicolumn{1}{c|}{} & 336 & \textbf{0.294} & \textbf{0.336} & 0.296 & 0.338 & 0.311 & 0.348 & 0.312 & 0.346 & 0.301 & 0.338 & 0.307 & 0.342 & 0.305 & 0.343 & 0.377 & 0.422 & 0.321 & 0.351 & 0.369 & 0.427 & 0.339 & 0.372 \\
\multicolumn{1}{c|}{} & 720 & \textbf{0.388} & \textbf{0.394} & 0.392 & \textbf{0.394} & 0.412 & 0.407 & 0.414 & 0.403 & 0.395 & 0.396 & 0.407 & 0.398 & 0.402 & 0.400 & 0.558 & 0.524 & 0.408 & 0.403 & 0.554 & 0.522 & 0.433 & 0.432 \\
\multicolumn{1}{c|}{} & Avg. & \textbf{0.272} & \textbf{0.320} & 0.274 & 0.322 & 0.288 & 0.332 & 0.288 & 0.330 & 0.280 & 0.325 & 0.286 & 0.327 & 0.281 & 0.326 & 0.358 & 0.404 & 0.291 & 0.333 & 0.350 & 0.401 & 0.327 & 0.371 \\ \midrule
\multicolumn{1}{c|}{\multirow{5}{*}{Traffic}} & 96 & 0.448 & 0.285 & 0.428 & 0.271 & \textbf{0.395} & \textbf{0.268} & 0.567 & 0.337 & 0.593 & 0.343 & 0.649 & 0.389 & 0.462 & 0.295 & 0.805 & 0.493 & 0.593 & 0.321 & 0.650 & 0.396 & 0.613 & 0.388 \\
\multicolumn{1}{c|}{} & 192 & 0.469 & 0.291 & 0.448 & 0.282 & \textbf{0.417} & \textbf{0.276} & 0.579 & 0.339 & 0.561 & 0.343 & 0.601 & 0.366 & 0.466 & 0.296 & 0.756 & 0.474 & 0.617 & 0.336 & 0.598 & 0.370 & 0.616 & 0.382 \\
\multicolumn{1}{c|}{} & 336 & 0.489 & 0.298 & 0.473 & 0.289 & \textbf{0.433} & \textbf{0.283} & 0.604 & 0.350 & 0.575 & 0.345 & 0.609 & 0.369 & 0.482 & 0.304 & 0.762 & 0.477 & 0.629 & 0.336 & 0.605 & 0.373 & 0.622 & 0.337 \\
\multicolumn{1}{c|}{} & 720 & 0.526 & 0.319 & 0.516 & 0.307 & \textbf{0.467} & \textbf{0.302} & 0.637 & 0.359 & 0.622 & 0.347 & 0.647 & 0.387 & 0.514 & 0.322 & 0.719 & 0.449 & 0.640 & 0.350 & 0.645 & 0.394 & 0.660 & 0.408 \\
\multicolumn{1}{c|}{} & Avg. & 0.483 & 0.298 & 0.466 & 0.287 & \textbf{0.428} & \textbf{0.282} & 0.597 & 0.346 & 0.588 & 0.344 & 0.626 & 0.378 & 0.481 & 0.304 & 0.760 & 0.473 & 0.620 & 0.336 & 0.625 & 0.383 & 0.628 & 0.379 \\ \midrule
\multicolumn{1}{c|}{\multirow{6}{*}{EPF}} & NP & \textbf{0.289} & \textbf{0.312} & 0.301 & 0.324 & 0.360 & 0.358 & 0.338 & 0.370 & 0.351 & 0.359 & 0.330 & 0.356 & 0.298 & 0.329 & 0.335 & 0.355 & 0.330 & 0.349 & 0.340 & 0.369 & 0.452 & 0.435 \\
\multicolumn{1}{c|}{} & PJM & \textbf{0.073} & \textbf{0.170} & 0.078 & 0.182 & 0.080 & 0.182 & 0.128 & 0.250 & 0.102 & 0.218 & 0.114 & 0.232 & 0.086 & 0.194 & 0.115 & 0.231 & 0.079 & 0.178 & 0.126 & 0.246 & 0.132 & 0.253 \\
\multicolumn{1}{c|}{} & BE & \textbf{0.145} & \textbf{0.168} & 0.148 & \textbf{0.168} & \textbf{0.145} & 0.172 & 0.172 & 0.212 & 0.173 & 0.206 & 0.182 & 0.215 & 0.155 & 0.176 & 0.193 & 0.229 & 0.151 & 0.176 & 0.200 & 0.236 & 0.195 & 0.225 \\
\multicolumn{1}{c|}{} & FR & 0.158 & 0.161 & 0.153 & 0.159 & \textbf{0.151} & \textbf{0.158} & 0.177 & 0.200 & 0.155 & 0.172 & 0.170 & 0.196 & 0.178 & 0.192 & 0.243 & 0.224 & 0.157 & 0.164 & 0.189 & 0.221 & 0.320 & 0.253 \\
\multicolumn{1}{c|}{} & DE & \textbf{0.193} & \textbf{0.241} & \textbf{0.193} & 0.246 & 0.211 & 0.264 & 0.238 & 0.309 & 0.231 & 0.280 & 0.229 & 0.292 & 0.198 & 0.253 & 0.231 & 0.292 & 0.201 & 0.258 & 0.235 & 0.304 & 0.324 & 0.359 \\
\multicolumn{1}{c|}{} & Avg. & \textbf{0.171} & \textbf{0.210} & 0.175 & 0.216 & 0.189 & 0.227 & 0.210 & 0.268 & 0.202 & 0.247 & 0.205 & 0.258 & 0.183 & 0.229 & 0.223 & 0.266 & 0.184 & 0.225 & 0.218 & 0.275 & 0.285 & 0.305 \\ \midrule
\multicolumn{2}{c|}{1\(^{\mathrm{st}}\) Count} & 31 & 28 & 8 & 8 & 7 & 9 & 0 & 0 & 0 & 2 & 0 & 1 & 0 & 1 & 0 & 0 & 0 & 0 & 0 & 0 & 0 & 0 \\ \bottomrule
\end{tabular}}
\end{table*}

Table~\ref{table: ms forecast} presents the results of time series forecasting with exogenous variables. CrossLinear incorporates a simple yet effective cross-correlation embedding mechanism combined with patching, which substantially enhances the performance of Linear-based models. As a result, CrossLinear outperforms the Transformer-based TimeXer on the majority of datasets, achieving top rankings in \textbf{30 cases for MSE and 29 cases for MAE}. It is worth noting that outliers in the Weather dataset were not preprocessed, which contributed to smaller error values~\cite{tsproblem}. However, for the sake of fair comparison, we follow the protocol of the previous study~\cite{timexer} and retain these outliers. In this setting, Linear-based models such as RLinear and TiDE demonstrate relatively better performance.

We also note that: (1) The CNN-based model, TimesNet, struggles to capture variable dependencies, leading to suboptimal performance, while the GNN-based MSGNet exhibits instability due to its intricate dependency modeling; (2) Among Transformer-based models, the two CD models, TimeXer and iTransformer, outperform the CI models, PatchTST and Autoformer. However, their performance may degrade due to excessive complexity in dependency modeling, which hampers generalization; (3) Linear-based models following the CI approach, such as SparseTSF, RLinear, and DLinear, exhibit weaker performance as they lack explicit mechanisms to model inter-variable dependencies; (4) Although TiDE incorporates variable dependencies, its performance is only marginally better than that of RLinear and DLinear, likely because its point-by-point time dependency modeling limits its ability to capture short-term patterns effectively.

As shown in Table~\ref{table: mm forecast}, in the multivariate time series forecasting setting, CrossLinear continues to achieve state-of-the-art results across most datasets, securing first-place rankings in \textbf{31 cases for MSE and 28 for MAE}. The iTransformer excels on the Traffic dataset due to its large number of exogenous variables, benefiting from the Transformer’s higher parameter capacity. Overall, CrossLinear demonstrates broad applicability and efficiency across forecasting tasks, aided by its weight-sharing mechanism.

\begin{table}[!htbp]
\caption{Ablation study. Results are averaged from all forecasting horizons. Full results can be found in Appendix \ref{sec: ablation study appendix}.}
\label{table: ablation}
\resizebox{1\linewidth}{!}{
\begin{tabular}{c|cc|cc|cc|cc}
\toprule
Dataset & \multicolumn{2}{c|}{ECL} & \multicolumn{2}{c|}{ETTh1} & \multicolumn{2}{c|}{Traffic} & \multicolumn{2}{c}{EPF} \\
Metric & MSE & MAE & MSE & MAE & MSE & MAE & MSE & MAE \\ \midrule
\textbf{Ours (Sum)} & \textbf{0.323} & \textbf{0.405} & \textbf{0.072} & \textbf{0.208} & \textbf{0.152} & \textbf{0.231} & \textbf{0.303} & \textbf{0.266} \\ \midrule
Endo only & 0.356 & 0.427 & 0.073 & 0.209 & 0.155 & 0.234 & 0.314 & 0.269 \\ \midrule
Cross only & 0.396 & 0.464 & 0.076 & 0.214 & 0.182 & 0.273 & 0.332 & 0.278 \\ \midrule
Concat & 0.327 & 0.413 & 0.076 & 0.214 & \textbf{0.152} & 0.235 & 0.315 & 0.272 \\ \bottomrule
\end{tabular}}
\end{table}

\subsection{Ablation Study}
\label{sec: ablation study}
CrossLinear utilizes the cross-correlation embedding technique to explicitly model the dependencies between variables. This is achieved by combining the endogenous variable with the cross-correlation embedding vector through a weighted summation. The weighting factor, denoted as $\alpha$, controls the relative contribution of the endogenous variable and the cross-correlation information. This integration strategy allows the model to benefit from both endogenous temporal patterns and variable interactions in a balanced and efficient manner.

To evaluate the effectiveness of this fusion mechanism and understand how each component contributes to the overall performance, we conducted an ablation study focusing on different ways of integrating the cross-correlation embedding. We designed three basic experimental settings: \textbf{Endo Only}, \textbf{Cross Only}, and \textbf{Concat}. Each setting isolates specific components of the input to assess their impacts.

In the Endo Only scenario, we set \(\alpha = 1\), which means only the endogenous variable is used as input while completely excluding the cross-correlation embedding. This allows us to establish a baseline performance based solely on endogenous time series patterns. In contrast, for the Cross Only case, we fix \(\alpha = 0\), using only the cross-correlation embedding vector as input. Although some endogenous information might still be captured through the cross-correlation process, this approach treats endogenous and exogenous variables equally, which can hinder the model’s ability to focus on critical endogenous dynamics. The third scenario, Concat, represents a different integration strategy where the endogenous variable and the cross-correlation embedding vector are concatenated rather than summed. This method preserves all original information without any blending, but it may introduce additional complexity due to increased dimensionality.

The full results of these ablation experiments are summarized in Table \ref{table: ablation}. These results show that our proposed summation-based integration method outperforms the other three strategies across multiple evaluation metrics. Interestingly, although mathematically, our method could be seen as equivalent to Cross Only under certain conditions, our approach achieves significantly better performance in practice. This indicates that treating endogenous and exogenous variables separately and then combining them in a controlled way leads to more effective learning compared to treating them equal.

One likely explanation for this outcome is that when endogenous and exogenous variables are given equal importance, especially in cases where the number of training samples is limited, the model may struggle to distinguish meaningful signals from noise or redundant features. As a result, the learned relationships become less generalizable and perform poorly during validation and testing. In contrast, our weighted summation approach enables the model to adaptively balance the influence of endogenous and exogenous information, leading to more robust forecasting outcomes.

\subsection{Generality Investigation}
\subsubsection{Cross-Correlation Embedding Generality}
The proposed cross-correlation embedding mechanism demonstrates a low level of integration complexity with existing forecasting models, which means it can be easily added to different existing models without requiring major modifications. This makes it a highly flexible and modular component. As a plug-and-play module, it can be applied across a wide range of time series forecasting frameworks. Its design is especially beneficial for models that follow a CI structure, as it enables the effective incorporation of both endogenous and exogenous variables into the model’s learning process. Importantly, this enhancement comes at a negligible increase in computational cost, making it a practical choice for real-world applications.

By integrating this module, models gain improved capability to handle forecasting tasks where exogenous variables play a role, thereby increasing their overall utility and adaptability. To validate its broad applicability, we implemented the cross-correlation embedding in five distinct forecasting models: SparseTSF, RLinear, PatchTST, DLinear, and Autoformer. In each case, the addition of the module led to measurable improvements in performance. Detailed results are summarized in Table \ref{table: embedding generality}, and full experimental outcomes can be found in Appendix \ref{sec: full results}.

\begin{table}[!htbp]
\caption{Performace promotion with our cross-correlation embedding. Full results can be found in Appendix \ref{sec: full results}.}
\label{table: embedding generality}
\resizebox{1\linewidth}{!}{
\begin{tabular}{cc|cc|cc|cc|cc|cc}
\toprule
\multicolumn{2}{c|}{\multirow{2}{*}{Model}} & \multicolumn{2}{c|}{SparseTSF} & \multicolumn{2}{c|}{RLinear} & \multicolumn{2}{c|}{PatchTST} & \multicolumn{2}{c|}{DLinear} & \multicolumn{2}{c}{Autoformer} \\
\multicolumn{2}{c|}{} & \multicolumn{2}{c|}{(2024)} & \multicolumn{2}{c|}{(2023)} & \multicolumn{2}{c|}{(2023)} & \multicolumn{2}{c|}{(2023)} & \multicolumn{2}{c}{(2021)} \\
\multicolumn{2}{c|}{Metric} & MSE & MAE & MSE & MAE & MSE & MAE & MSE & MAE & MSE & MAE \\ \midrule
\multicolumn{1}{c|}{\multirow{2}{*}{ECL}} & Ori. & 0.372 & 0.431 & 0.444 & 0.486 & 0.394 & 0.446 & 0.393 & \textbf{0.457} & 0.495 & 0.528 \\
\multicolumn{1}{c|}{} & \textbf{+ Emb.} & \textbf{0.350} & \textbf{0.420} & \textbf{0.406} & \textbf{0.473} & \textbf{0.359} & \textbf{0.427} & \textbf{0.383} & 0.463 & \textbf{0.469} & \textbf{0.514} \\ \midrule
\multicolumn{1}{c|}{\multirow{2}{*}{ETTh1}} & Ori. & 0.079 & 0.224 & 0.084 & 0.224 & 0.078 & 0.215 & 0.116 & 0.259 & 0.130 & 0.282 \\
\multicolumn{1}{c|}{} & \textbf{+ Emb.} & \textbf{0.078} & \textbf{0.221} & \textbf{0.080} & \textbf{0.218} & \textbf{0.077} & \textbf{0.214} & \textbf{0.113} & \textbf{0.256} & \textbf{0.114} & \textbf{0.264} \\ \midrule
\multicolumn{1}{c|}{\multirow{2}{*}{Traffic}} & Ori. & 0.194 & 0.274 & 0.324 & 0.412 & 0.173 & 0.253 & 0.323 & 0.404 & 0.293 & 0.392 \\
\multicolumn{1}{c|}{} & \textbf{+ Emb.} & \textbf{0.181} & \textbf{0.268} & \textbf{0.234} & \textbf{0.325} & \textbf{0.166} & \textbf{0.249} & \textbf{0.308} & \textbf{0.391} & \textbf{0.270} & \textbf{0.367} \\ \midrule
\multicolumn{1}{c|}{\multirow{2}{*}{EPF}} & Ori. & 0.351 & 0.305 & 0.412 & 0.339 & 0.330 & \textbf{0.282} & 0.366 & 0.314 & 0.453 & 0.368 \\
\multicolumn{1}{c|}{} & \textbf{+ Emb.} & \textbf{0.343} & \textbf{0.301} & \textbf{0.348} & \textbf{0.304} & \textbf{0.328} & \textbf{0.282} & \textbf{0.364} & \textbf{0.313} & \textbf{0.426} & \textbf{0.352} \\ \bottomrule
\end{tabular}}
\end{table}

The evaluation results indicate that for most of the tested models, the inclusion of the cross-correlation embedding leads to better forecasting accuracy across multiple datasets. For instance, when applied to the SparseTSF model, the MSE was reduced by 5.9\%, 1.3\%, 5.2\%, and 2.8\% on the ECL, ETTh1, Traffic, and EPF datasets, respectively. Similarly, the RLinear model showed even more significant improvements, with MSE reductions of 8.6\%, 4.8\%, 27.8\%, and 15.5\% across the same four datasets. Particularly notable is the 27.8\% reduction in error on the Traffic dataset, demonstrating the strong impact of the module in handling complex relationships between exogenous variables and the target time series.

It can be found that the ECL and Traffic datasets contain a large number of exogenous variables, with 320 and 861, respectively, while the ETTh1 dataset has only 6 such variables. The observed performance gains are more pronounced in datasets with a higher number of exogenous variables, suggesting that the cross-correlation embedding can benefit from a richer exogenous information. This insight highlights the module’s potential in real-world forecasting scenarios where exogenous data sources are abundant and informative.

\subsubsection{Missing Values}
In complex real-world scenarios, historical time series data are often incomplete due to uncontrollable factors such as sensor failures, making robust forecasting models essential. To further assess the generalization ability of CrossLinear under such conditions, we conducted experiments simulating scenarios with missing historical time series data. Specifically, we simulate missing values by randomly masking the endogenous or exogenous variables, replacing them with either zeros or random values sampled from the normal distribution \(\mathcal{N}(0,\ 1)\), in order to assess the model’s robustness to incomplete information. The corresponding results are reported in Table~\ref{table: missing values}.

\begin{table}[!htbp]
\caption{Performance under missing values.}
\label{table: missing values}
\resizebox{1\linewidth}{!}{
\begin{tabular}{c|c|c|cc|cc|cc|cc}
\toprule
\multicolumn{3}{c|}{} & \multicolumn{2}{c|}{ECL} & \multicolumn{2}{c|}{ETTh1} & \multicolumn{2}{c|}{Traffic} & \multicolumn{2}{c}{NP} \\
\multicolumn{3}{c|}{} & \multicolumn{2}{c|}{(96-to-96)} & \multicolumn{2}{c|}{(96-to-96)} & \multicolumn{2}{c|}{(96-to-96)} & \multicolumn{2}{c}{(96-to-96)} \\
\multicolumn{3}{c|}{\multirow{-3}{*}{Masking   Method}} & MSE & MAE & MSE & MAE & MSE & MAE & MSE & MAE \\ \midrule
 &  & 50\%-Mask & 0.292 & 0.391 & 0.057 & 0.181 & 0.158 & 0.235 & 0.292 & 0.321 \\
 & \multirow{-2}{*}{Endo} & 100\%-Mask & 0.448 & 0.506 & 0.066 & 0.205 & 0.233 & 0.336 & 0.331 & 0.348 \\ \cmidrule{2-11} 
 &  & 50\%-Mask & 0.255 & 0.361 & 0.055 & 0.179 & \textbf{0.142} & \textbf{0.214} & 0.238 & 0.270 \\
\multirow{-4}{*}{Zero} & \multirow{-2}{*}{Exo} & 100\%-Mask & 0.322 & 0.407 & 0.055 & 0.180 & 0.208 & 0.302 & 0.240 & 0.270 \\ \midrule
 &  & 50\%-Mask & 0.322 & 0.418 & 0.058 & 0.184 & 0.167 & 0.243 & 0.327 & 0.346 \\
 & \multirow{-2}{*}{Endo} & 100\%-Mask & 0.354 & 0.441 & 0.063 & 0.193 & 0.170 & 0.246 & 0.334 & 0.353 \\ \cmidrule{2-11} 
 &  & 50\%-Mask & 0.269 & 0.370 & 0.055 & 0.179 & 0.149 & 0.222 & 0.240 & 0.272 \\
\multirow{-4}{*}{Random} & \multirow{-2}{*}{Exo} & 100\%-Mask & 0.299 & 0.388 & 0.056 & 0.180 & 0.226 & 0.323 & 0.243 & 0.275 \\ \midrule
\multicolumn{3}{c|}{No Masking} & \textbf{0.251} & \textbf{0.359} & \textbf{0.055} & \textbf{0.178} & 0.149 & 0.223 & \textbf{0.232} & \textbf{0.268} \\ \bottomrule
\end{tabular}}
\end{table}

An interesting observation is that a certain amount of missing exogenous data in the Traffic dataset slightly improves performance, likely because the random masking acts as a form of data augmentation. In the ETTh1 dataset, the impact of missing exogenous variables is minimal due to their limited quantity (only 6 exogenous variables), even when all are masked.

Overall, as the proportion of missing data increases, model performance tends to degrade. However, even when all endogenous data is absent, the model can still utilize exogenous variable information through the cross-correlation embedding, maintaining errors within a manageable range. Additionally, when exogenous variables are entirely missing, the model’s performance declines significantly, emphasizing the crucial role of exogenous variable integration in achieving precise forecasts.

\subsection{Model Analysis}
\subsubsection{Hyperparameter Sensitivity}
\begin{figure}[!htbp]
\centering
\includegraphics[width=1\linewidth]{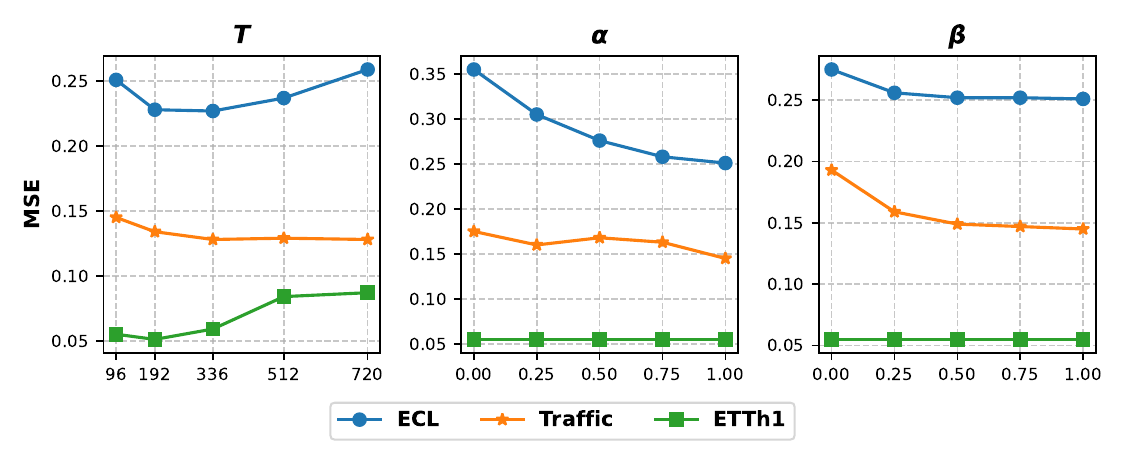}
\caption{Hyperparameter sensitivity with respect to \(T\), \(\alpha\), and \(\beta\). For \(T\), \(S=96\); for \(\alpha\) and \(\beta\), \(T=96\) and \(S=96\). More results can be found in Appendix \ref{sec: hyperparameter sensitivity}.}
\label{fig: Param Analysis}
\end{figure}

To investigate the impact of different hyperparameter settings on model performance, this section evaluates the effects of three hyperparameters: the length of the lookback window (\(T\)), \(\alpha\), and \(\beta\) (see Figure \ref{fig: Param Analysis}). For additional hyperparameter analysis, please refer to Appendix \ref{sec: hyperparameter sensitivity}.

It was found that a larger lookback window does not necessarily improve results. Datasets with more exogenous variables tend to benefit more from larger windows, aligning with current research~\cite{scalinglaw,itransformer}. From a practical standpoint, the historical state of a system from long ago is less likely to influence the present or predict the future. Unlike natural language processing, long-term temporal dependencies are generally weaker in time series analysis. How to use the dependencies between various variables to assist forecasting tasks is a focus in academia~\cite{crossformer,timexer}.

Additionally, the analysis of \(\alpha\) and \(\beta\) shows better performance when initialized closer to 1. This may be due to training dynamics, where larger initial values of \(\alpha\) and \(\beta\) prioritize the endogenous variable in early training stages, leading to a more stable optimization process and improved convergence and performance.

\subsubsection{Variate-wise Correlation Analysis}
\begin{figure}[!htbp]
\centering
\includegraphics[width=1\linewidth]{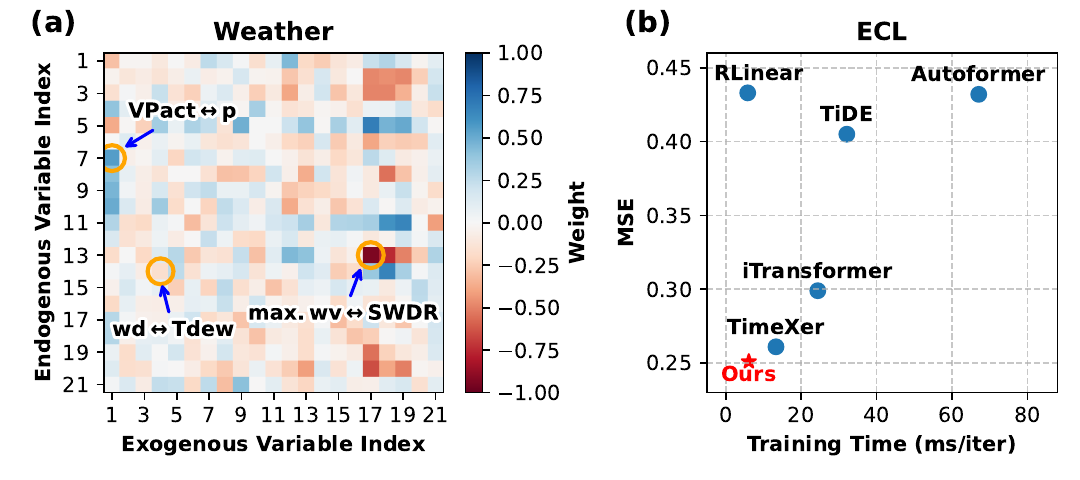}
\caption{Model anlysis. (a) Variate-wise correlation analysis. (b) Model efficiency analysis.}
\label{fig: Correlation Efficiency}
\end{figure}

CrossLinear employs 1D convolution to capture variable dependencies, with the convolutional layer’s weights acting as indicators of these relationships. In the context of multivariate time series prediction, which includes \(N\) variables, the weight matrix, denoted as \(\mathrm{MatWeigh} \in \mathbb{R}^{N \times N \times 3}\) (where 3 represents the kernel size), encapsulates these dependencies. To facilitate visualization, we aggregate the weights along the last dimension, allowing the resulting values at row \(i\) and column \(j\) to roughly represent the dependency between the \(i^{\mathrm{th}}\) and \(j^{\mathrm{th}}\) variables, as shown in Figure \ref{fig: Correlation Efficiency}(a).

In the figure, blue regions indicate positive dependencies, red regions represent negative dependencies, and the color intensity reflects the strength of the relationship. For example, actual vapor pressure (VPact) exhibits a positive correlation on air pressure (p), while maximum wind speed (max. wv) shows a notable negative dependency on shortwave downward radiation (SWDR). Conversely, no significant relationship is observed between wind direction (wd) and dew point temperature (Tdew). These findings are consistent with current research~\cite{interpretability_1,interpretability_2}, and validate the efficacy of CrossLinear in identifying variate-wise dependencies.

\subsubsection{Model Efficiency Analysis}
The complexity of the proposed cross-correlation embedding is \(O(T)\), the complexity of the patch embedding is \(O(T/p)\), and the complexity of the head is also \(O(T)\). Therefore, the overall complexity of the model remains at \(O(T)\). A comparison of this complexity with other methods is shown in Table~\ref{table: complexity comparison}. To evaluate the efficiency of CrossLinear, we compared it with six baseline models under the same batch size. As shown in Figure \ref{fig: Correlation Efficiency}(b), Autoformer, which is not specifically designed for many-to-one forecasting, demonstrates low efficiency. Although TiDE is proposed for such tasks, its code is not optimized for this purpose, leading to similarly low efficiency. In contrast, CrossLinear and TimeXer incorporate targeted design strategies to improve efficiency. Notably, CrossLinear outperforms TimeXer, with each iteration taking less time while achieving better precision. Overall, CrossLinear strikes an optimal balance between training time and error, significantly enhancing the forecasting performance of Linear-based models and surpassing Transformer-based models.

\begin{table}[!htbp]
\caption{Computational complexity comparison.}
\label{table: complexity comparison}
\resizebox{1\linewidth}{!}{
\begin{tabular}{c|c|c|c|c|c}
\toprule
Model & \textbf{CrossLinear} & TimeXer & RLinear & PatchTST & Autoformer \\
 & \textbf{(Ours)} & (2024) & (2023) & (2023) & (2021) \\ \midrule
Complexity & \(O(T)\) & \(O((T/p)^2)\) & \(O(T)\) & \(O((T/p)^2)\) & \(O(T\log T)\) \\ \bottomrule
\end{tabular}}
\end{table}

\section{Conclusions}
In this paper, we introduce \textbf{CrossLinear}, an advanced model specifically designed to improve time series forecasting with exogenous variables. The core innovation of CrossLinear lies in its unique architecture, which is a flexible plug-and-play cross-correlation embedding module. This module enables the model to effectively capture variable dependencies. Also, CrossLinear adopt patching technique to capture temporal dependencies. Through a series of comprehensive experiments conducted on multiple real-world datasets, we demonstrate that CrossLinear outperforms existing methods in most cases.

One of the key advantages of CrossLinear is its lightweight structure, which ensures high computational efficiency and makes it suitable for deployment in resource-constrained environments. This design also enhances the model’s adaptability, allowing it to be easily integrated into different forecasting frameworks.

Although CrossLinear achieves impressive results in time series forecasting, there are still opportunities for further work. One direction is to explore its applicability to tasks beyond time series forecasting, such as anomaly detection or classification. Another promising research direction is the integration of diffusion models into our framework.

\section*{Acknowledgements}
The research was partially supported by ``Pioneer'' and ``Leading Goose'' R\&D Program of Zhejiang 2023C01029, the China National Natural Science Foundation with No. 62132018.

\clearpage
\bibliographystyle{ACM-Reference-Format}
\bibliography{reference}

\clearpage
\appendix

\section{Details of Experiments}
\label{sec: details of experiments}
All experiments are implemented using the PyTorch framework~\cite{pytorch} and executed on a single NVIDIA GeForce RTX 4090 GPU equipped with 24GB of memory. We utilize the Adam optimizer~\cite{adam} for model training, with the initial learning rate selected from the range \([10^{-5},\ 10^{-3}]\). Model optimization is performed using the L2 loss function. The batch size is varied within the range \([8,\ 64]\), and the number of training epochs is fixed at 10 across all experiments to ensure consistency.

In our proposed model, the hyperparameter \(\alpha\) is chosen from the interval \([0, 1]\), while \(\beta\) is also selected from \([0, 1]\). The kernel size and stride in the cross-correlation embedding module are fixed at \(3\) and \(1\), respectively.

For comparative evaluation, we benchmark our method against several strong baselines, including TimeXer~\cite{timexer}, MSGNet~\cite{msgnet}, and SparseTSF~\cite{sparsetsf}. All baseline implementations strictly follow the configurations and settings provided in their respective original papers or official repositories to ensure fair and reproducible comparisons.

\section{Ablation Study}
\label{sec: ablation study appendix}
CrossLinear leverages cross-correlation embedding to explicitly capture inter-variable dependencies by computing a weighted summation of the endogenous variable and the cross-correlation embedding vector, thereby enabling a more comprehensive representation of temporal interactions. To rigorously evaluate the effectiveness of this embedding integration strategy, we conducted a detailed ablation study focusing on the embedding module. The complete results of this analysis, which compare performance on different embedding techniques, are summarized in Tables~\ref{table: full long ablation} and~\ref{table: full short ablation}.

\begin{table}[!htbp]
\caption{Full results of ablation study on long-term many-to-one forecasting. ``Avg.'' refers to the average results across all datasets. The best results are highlighted in bold.}
\label{table: full long ablation}
\resizebox{1\linewidth}{!}{
\begin{tabular}{c|c|cc|cc|cc}
\toprule
\multirow{2}{*}{Design} & \multirow{2}{*}{Horizon} & \multicolumn{2}{c|}{ECL} & \multicolumn{2}{c|}{ETTh1} & \multicolumn{2}{c}{Traffic} \\
 &  & MSE & MAE & MSE & MAE & MSE & MAE \\ \midrule
\textbf{Ours (Sum)} & 96 & 0.251 & 0.359 & 0.055 & 0.178 & 0.149 & 0.223 \\
 & 192 & 0.294 & 0.381 & 0.072 & 0.205 & 0.149 & 0.225 \\
 & 336 & 0.343 & 0.416 & 0.082 & 0.226 & 0.148 & 0.229 \\
 & 720 & 0.403 & 0.465 & 0.080 & 0.225 & 0.161 & 0.247 \\
 & Avg. & \textbf{0.323} & \textbf{0.405} & \textbf{0.072} & \textbf{0.208} & \textbf{0.152} & \textbf{0.231} \\ \midrule
Endo only & 96 & 0.290 & 0.381 & 0.055 & 0.179 & 0.150 & 0.229 \\
 & 192 & 0.326 & 0.406 & 0.072 & 0.205 & 0.150 & 0.227 \\
 & 336 & 0.373 & 0.435 & 0.084 & 0.227 & 0.149 & 0.231 \\
 & 720 & 0.436 & 0.484 & 0.080 & 0.225 & 0.170 & 0.251 \\
 & Avg. & 0.356 & 0.427 & 0.073 & 0.209 & 0.155 & 0.234 \\ \midrule
Cross only & 96 & 0.357 & 0.447 & 0.055 & 0.178 & 0.175 & 0.268 \\
 & 192 & 0.356 & 0.436 & 0.072 & 0.207 & 0.177 & 0.266 \\
 & 336 & 0.414 & 0.468 & 0.082 & 0.226 & 0.183 & 0.277 \\
 & 720 & 0.455 & 0.505 & 0.096 & 0.245 & 0.191 & 0.281 \\
 & Avg. & 0.396 & 0.464 & 0.076 & 0.214 & 0.182 & 0.273 \\ \midrule
Concat & 96 & 0.271 & 0.376 & 0.057 & 0.182 & 0.145 & 0.226 \\
 & 192 & 0.301 & 0.393 & 0.073 & 0.208 & 0.147 & 0.227 \\
 & 336 & 0.352 & 0.430 & 0.085 & 0.229 & 0.149 & 0.236 \\
 & 720 & 0.384 & 0.453 & 0.090 & 0.238 & 0.165 & 0.250 \\
 & Avg. & 0.327 & 0.413 & 0.076 & 0.214 & \textbf{0.152} & 0.235 \\ \bottomrule
\end{tabular}}
\end{table}

\begin{table}[!htbp]
\caption{Full results of ablation study on short-term many-to-one forecasting. ``Avg.'' refers to the average results across all datasets. The best results are highlighted in bold.}
\label{table: full short ablation}
\resizebox{0.9\linewidth}{!}{
\begin{tabular}{cc|c|c|c|c}
\toprule
\multicolumn{2}{c|}{Design} & \textbf{Ours (Sum)} & Endo only & Cross only & Concat \\ \midrule
\multirow{2}{*}{NP} & MSE & \textbf{0.232} & 0.240 & 0.239 & 0.243 \\
 & MAE & \textbf{0.268} & 0.270 & 0.274 & 0.276 \\ \midrule
\multirow{2}{*}{PJM} & MSE & \textbf{0.092} & 0.094 & 0.094 & 0.097 \\
 & MAE & \textbf{0.193} & 0.194 & 0.197 & 0.197 \\ \midrule
\multirow{2}{*}{BE} & MSE & 0.372 & 0.380 & \textbf{0.371} & 0.378 \\
 & MAE & \textbf{0.248} & 0.251 & 0.251 & 0.252 \\ \midrule
\multirow{2}{*}{FR} & MSE & \textbf{0.384} & 0.387 & 0.512 & 0.389 \\
 & MAE & 0.207 & \textbf{0.206} & 0.251 & 0.208 \\ \midrule
\multirow{2}{*}{DE} & MSE & \textbf{0.437} & 0.469 & 0.446 & 0.468 \\
 & MAE & \textbf{0.413} & 0.426 & 0.418 & 0.423 \\ \midrule
\multirow{2}{*}{Avg.} & MSE & \textbf{0.303} & 0.314 & 0.332 & 0.315 \\
 & MAE & \textbf{0.266} & 0.269 & 0.278 & 0.272 \\ \bottomrule
\end{tabular}}
\end{table}

The abovementioned ablation study primarily investigates the effectiveness of the cross-correlation embedding. In addition, we conducted a separate ablation study on the impact of positional embedding, with the results presented in Table~\ref{table: ablation study on pe}.

\begin{table}[!htbp]
\caption{Ablation study on positional embedding. ``Avg.'' refers to the average results across all five datasets. The best results are highlighted in bold.}
\label{table: ablation study on pe}
\resizebox{1\linewidth}{!}{
\begin{tabular}{l|l|ll|ll|ll}
\toprule
\multirow{2}{*}{Design} & \multirow{2}{*}{Horizon} & \multicolumn{2}{l|}{ECL} & \multicolumn{2}{l|}{ETTh1} & \multicolumn{2}{l}{Traffic} \\
 &  & MSE & MAE & MSE & MAE & MSE & MAE \\ \midrule
\multirow{5}{*}{w/ PE} & 96 & 0.251 & 0.359 & 0.055 & 0.178 & 0.149 & 0.223 \\
 & 192 & 0.294 & 0.381 & 0.072 & 0.205 & 0.149 & 0.225 \\
 & 336 & 0.343 & 0.416 & 0.082 & 0.226 & 0.148 & 0.229 \\
 & 720 & 0.403 & 0.465 & 0.080 & 0.225 & 0.161 & 0.247 \\
 & Avg & \textbf{0.323} & \textbf{0.405} & \textbf{0.072} & \textbf{0.208} & \textbf{0.152} & \textbf{0.231} \\ \midrule
\multirow{5}{*}{w/o PE} & 96 & 0.248 & 0.357 & 0.055 & 0.178 & 0.155 & 0.233 \\
 & 192 & 0.303 & 0.390 & 0.072 & 0.205 & 0.152 & 0.230 \\
 & 336 & 0.349 & 0.423 & 0.083 & 0.226 & 0.151 & 0.232 \\
 & 720 & 0.406 & 0.467 & 0.085 & 0.230 & 0.167 & 0.254 \\
 & Avg & 0.327 & 0.409 & 0.074 & 0.210 & 0.156 & 0.237 \\ \bottomrule
\end{tabular}}
\end{table}

Mathematically, our method can be shown to be equivalent to Cross Only's approach. Proof can be found below. In fact, 1D convolution can be seen as attention (or graph neural network) with some specific prior knowledge.

\begin{table*}[!htbp]
\caption{Full results of performace promotion with our cross-correlation embedding on multiple models. ``Avg.'' refers to the average results across all datasets. The best results are highlighted in bold.}
\label{table: full embedding generality}
\resizebox{0.77\linewidth}{!}{
\begin{tabular}{ccc|cc|cc|cc|cc|cc}
\toprule
\multicolumn{3}{c|}{\multirow{2}{*}{Model}} & \multicolumn{2}{c|}{SparseTSF} & \multicolumn{2}{c|}{RLinear} & \multicolumn{2}{c|}{PatchTST} & \multicolumn{2}{c|}{DLinear} & \multicolumn{2}{c}{Autoformer} \\
\multicolumn{3}{c|}{} & \multicolumn{2}{c|}{(2024)} & \multicolumn{2}{c|}{(2023)} & \multicolumn{2}{c|}{(2023)} & \multicolumn{2}{c|}{(2023)} & \multicolumn{2}{c}{(2021)} \\
\multicolumn{3}{c|}{Metric} & MSE & MAE & MSE & MAE & MSE & MAE & MSE & MAE & MSE & MAE \\ \midrule
\multicolumn{1}{c|}{\multirow{10}{*}{ECL}} & \multicolumn{1}{c|}{\multirow{5}{*}{Ori.}} & 96 & 0.332 & 0.404 & 0.433 & 0.480 & 0.339 & 0.412 & 0.387 & 0.451 & 0.432 & 0.502 \\
\multicolumn{1}{c|}{} & \multicolumn{1}{c|}{} & 192 & 0.335 & 0.402 & 0.407 & 0.461 & 0.361 & 0.425 & 0.365 & 0.436 & 0.492 & 0.492 \\
\multicolumn{1}{c|}{} & \multicolumn{1}{c|}{} & 336 & 0.378 & 0.432 & 0.440 & 0.481 & 0.393 & 0.440 & 0.391 & 0.453 & 0.508 & 0.548 \\
\multicolumn{1}{c|}{} & \multicolumn{1}{c|}{} & 720 & 0.444 & 0.486 & 0.495 & 0.523 & 0.482 & 0.507 & 0.428 & 0.481 & 0.547 & 0.569 \\
\multicolumn{1}{c|}{} & \multicolumn{1}{c|}{} & Avg. & 0.372 & 0.431 & 0.444 & 0.486 & 0.394 & 0.446 & 0.393 & \textbf{0.457} & 0.495 & 0.528 \\ \cmidrule{2-13} 
\multicolumn{1}{c|}{} & \multicolumn{1}{c|}{\multirow{5}{*}{\textbf{+ Emb.}}} & 96 & 0.300 & 0.387 & 0.377 & 0.460 & 0.275 & 0.375 & 0.384 & 0.463 & 0.430 & 0.498 \\
\multicolumn{1}{c|}{} & \multicolumn{1}{c|}{} & 192 & 0.309 & 0.398 & 0.376 & 0.451 & 0.339 & 0.413 & 0.355 & 0.446 & 0.496 & 0.520 \\
\multicolumn{1}{c|}{} & \multicolumn{1}{c|}{} & 336 & 0.363 & 0.422 & 0.416 & 0.475 & 0.367 & 0.430 & 0.379 & 0.460 & 0.455 & 0.506 \\
\multicolumn{1}{c|}{} & \multicolumn{1}{c|}{} & 720 & 0.426 & 0.475 & 0.456 & 0.508 & 0.454 & 0.490 & 0.414 & 0.484 & 0.497 & 0.530 \\
\multicolumn{1}{c|}{} & \multicolumn{1}{c|}{} & Avg. & \textbf{0.350} & \textbf{0.420} & \textbf{0.406} & \textbf{0.473} & \textbf{0.359} & \textbf{0.427} & \textbf{0.383} & 0.463 & \textbf{0.469} & \textbf{0.514} \\ \midrule
\multicolumn{1}{c|}{\multirow{10}{*}{ETTh1}} & \multicolumn{1}{c|}{\multirow{5}{*}{Ori.}} & 96 & 0.063 & 0.199 & 0.059 & 0.185 & 0.055 & 0.178 & 0.065 & 0.188 & 0.119 & 0.263 \\
\multicolumn{1}{c|}{} & \multicolumn{1}{c|}{} & 192 & 0.080 & 0.224 & 0.078 & 0.214 & 0.072 & 0.206 & 0.088 & 0.222 & 0.132 & 0.286 \\
\multicolumn{1}{c|}{} & \multicolumn{1}{c|}{} & 336 & 0.091 & 0.241 & 0.093 & 0.240 & 0.087 & 0.231 & 0.110 & 0.257 & 0.126 & 0.278 \\
\multicolumn{1}{c|}{} & \multicolumn{1}{c|}{} & 720 & 0.081 & 0.229 & 0.106 & 0.256 & 0.098 & 0.247 & 0.202 & 0.371 & 0.143 & 0.299 \\
\multicolumn{1}{c|}{} & \multicolumn{1}{c|}{} & Avg. & 0.079 & 0.224 & 0.084 & 0.224 & 0.078 & 0.215 & 0.116 & 0.259 & 0.130 & 0.282 \\ \cmidrule{2-13} 
\multicolumn{1}{c|}{} & \multicolumn{1}{c|}{\multirow{5}{*}{\textbf{+ Emb.}}} & 96 & 0.059 & 0.191 & 0.056 & 0.179 & 0.056 & 0.179 & 0.064 & 0.188 & 0.086 & 0.230 \\
\multicolumn{1}{c|}{} & \multicolumn{1}{c|}{} & 192 & 0.082 & 0.227 & 0.076 & 0.210 & 0.073 & 0.208 & 0.087 & 0.221 & 0.116 & 0.264 \\
\multicolumn{1}{c|}{} & \multicolumn{1}{c|}{} & 336 & 0.090 & 0.239 & 0.091 & 0.237 & 0.087 & 0.231 & 0.106 & 0.253 & 0.120 & 0.272 \\
\multicolumn{1}{c|}{} & \multicolumn{1}{c|}{} & 720 & 0.079 & 0.227 & 0.097 & 0.245 & 0.092 & 0.240 & 0.194 & 0.363 & 0.134 & 0.291 \\
\multicolumn{1}{c|}{} & \multicolumn{1}{c|}{} & Avg. & \textbf{0.078} & \textbf{0.221} & \textbf{0.080} & \textbf{0.218} & \textbf{0.077} & \textbf{0.214} & \textbf{0.113} & \textbf{0.256} & \textbf{0.114} & \textbf{0.264} \\ \midrule
\multicolumn{1}{c|}{\multirow{10}{*}{Traffic}} & \multicolumn{1}{c|}{\multirow{5}{*}{Ori.}} & 96 & 0.204 & 0.288 & 0.350 & 0.431 & 0.176 & 0.253 & 0.268 & 0.351 & 0.250 & 0.343 \\
\multicolumn{1}{c|}{} & \multicolumn{1}{c|}{} & 192 & 0.187 & 0.267 & 0.314 & 0.404 & 0.162 & 0.243 & 0.302 & 0.387 & 0.294 & 0.396 \\
\multicolumn{1}{c|}{} & \multicolumn{1}{c|}{} & 336 & 0.186 & 0.264 & 0.305 & 0.399 & 0.164 & 0.248 & 0.298 & 0.384 & 0.322 & 0.416 \\
\multicolumn{1}{c|}{} & \multicolumn{1}{c|}{} & 720 & 0.197 & 0.279 & 0.328 & 0.415 & 0.189 & 0.267 & 0.340 & 0.416 & 0.307 & 0.414 \\
\multicolumn{1}{c|}{} & \multicolumn{1}{c|}{} & Avg. & 0.194 & 0.274 & 0.324 & 0.412 & 0.173 & 0.253 & 0.323 & 0.404 & 0.293 & 0.392 \\ \cmidrule{2-13} 
\multicolumn{1}{c|}{} & \multicolumn{1}{c|}{\multirow{5}{*}{\textbf{+ Emb.}}} & 96 & 0.176 & 0.265 & 0.268 & 0.359 & 0.159 & 0.238 & 0.338 & 0.414 & 0.269 & 0.364 \\
\multicolumn{1}{c|}{} & \multicolumn{1}{c|}{} & 192 & 0.183 & 0.272 & 0.222 & 0.311 & 0.162 & 0.242 & 0.260 & 0.356 & 0.276 & 0.372 \\
\multicolumn{1}{c|}{} & \multicolumn{1}{c|}{} & 336 & 0.178 & 0.267 & 0.222 & 0.314 & 0.163 & 0.249 & 0.285 & 0.378 & 0.266 & 0.365 \\
\multicolumn{1}{c|}{} & \multicolumn{1}{c|}{} & 720 & 0.188 & 0.269 & 0.224 & 0.315 & 0.182 & 0.267 & 0.349 & 0.417 & 0.269 & 0.367 \\
\multicolumn{1}{c|}{} & \multicolumn{1}{c|}{} & Avg. & \textbf{0.181} & \textbf{0.268} & \textbf{0.234} & \textbf{0.325} & \textbf{0.166} & \textbf{0.249} & \textbf{0.308} & \textbf{0.391} & \textbf{0.270} & \textbf{0.367} \\ \midrule
\multicolumn{1}{c|}{\multirow{12}{*}{EPF}} & \multicolumn{1}{c|}{\multirow{6}{*}{Ori.}} & NP & 0.310 & 0.323 & 0.335 & 0.340 & 0.267 & 0.284 & 0.309 & 0.321 & 0.402 & 0.398 \\
\multicolumn{1}{c|}{} & \multicolumn{1}{c|}{} & PJM & 0.115 & 0.229 & 0.124 & 0.229 & 0.106 & 0.209 & 0.108 & 0.215 & 0.168 & 0.267 \\
\multicolumn{1}{c|}{} & \multicolumn{1}{c|}{} & BE & 0.432 & 0.289 & 0.520 & 0.337 & 0.400 & 0.262 & 0.463 & 0.313 & 0.500 & 0.333 \\
\multicolumn{1}{c|}{} & \multicolumn{1}{c|}{} & FR & 0.384 & 0.223 & 0.507 & 0.290 & 0.411 & 0.220 & 0.429 & 0.260 & 0.519 & 0.295 \\
\multicolumn{1}{c|}{} & \multicolumn{1}{c|}{} & DE & 0.513 & 0.464 & 0.574 & 0.498 & 0.461 & 0.432 & 0.520 & 0.463 & 0.674 & 0.544 \\
\multicolumn{1}{c|}{} & \multicolumn{1}{c|}{} & Avg. & 0.351 & 0.305 & 0.412 & 0.339 & 0.330 & \textbf{0.282} & 0.366 & 0.314 & 0.453 & 0.368 \\ \cmidrule{2-13} 
\multicolumn{1}{c|}{} & \multicolumn{1}{c|}{\multirow{6}{*}{\textbf{+ Emb.}}} & NP & 0.302 & 0.319 & 0.293 & 0.310 & 0.258 & 0.285 & 0.309 & 0.321 & 0.365 & 0.368 \\
\multicolumn{1}{c|}{} & \multicolumn{1}{c|}{} & PJM & 0.113 & 0.227 & 0.108 & 0.215 & 0.107 & 0.208 & 0.107 & 0.213 & 0.146 & 0.254 \\
\multicolumn{1}{c|}{} & \multicolumn{1}{c|}{} & BE & 0.424 & 0.281 & 0.453 & 0.302 & 0.396 & 0.263 & 0.464 & 0.315 & 0.512 & 0.329 \\
\multicolumn{1}{c|}{} & \multicolumn{1}{c|}{} & FR & 0.372 & 0.218 & 0.402 & 0.245 & 0.406 & 0.219 & 0.426 & 0.258 & 0.493 & 0.286 \\
\multicolumn{1}{c|}{} & \multicolumn{1}{c|}{} & DE & 0.503 & 0.460 & 0.485 & 0.448 & 0.472 & 0.436 & 0.513 & 0.460 & 0.614 & 0.521 \\
\multicolumn{1}{c|}{} & \multicolumn{1}{c|}{} & Avg. & \textbf{0.343} & \textbf{0.301} & \textbf{0.348} & \textbf{0.304} & \textbf{0.328} & \textbf{0.282} & \textbf{0.364} & \textbf{0.313} & \textbf{0.426} & \textbf{0.352} \\ \bottomrule
\end{tabular}}
\end{table*}

\textbf{Proof.} To prove the mathematical equivalence between the Summation method (ours) and the Cross Only method, we need to demonstrate that their final embedding vectors are identical. In the Summation method, the final embedding vector is computed as a weighted sum of the endogenous variable and the result of a convolution operation on the concatenated endogenous and exogenous variables:

\begin{equation}
\alpha \cdot \mathbf{X}_{1:T,1}^{endo*} + (1-\alpha) \cdot \mathrm{Conv1D}(\mathrm{Stack}(\mathbf{X}_{1:T,1}^{endo*},\ \mathbf{X}_{1:T,N-1}^{exo*}))
\end{equation}

In the Cross Only method, the embedding vector is computed by applying the convolution operation on the concatenated endogenous and exogenous variables:

\begin{equation}
\mathrm{Conv1D^{\prime}}(\mathrm{Stack}(\mathbf{X}_{1:T,1}^{endo*},\ \mathbf{X}_{1:T,N-1}^{exo*}))
\end{equation}

To establish the equivalence of these two methods, we need to construct the kernel \(K\) of the Summation method and the kernel \(K^{\prime}\) for the Cross Only method, ensuring that when applied to the same input, the outputs from both methods are identical. We can define a selection matrix \(S\), which retains only the endogenous variable \(\mathbf{X}_{1:T,1}^{endo*}\) and sets the exogenous variables \(\mathbf{X}_{1:T,N-1}^{exo*}\) to zero. Specifically, the matrix \(S\) ensures that:

\begin{equation}
S * \mathrm{Stack}(\mathbf{X}_{1:T,1}^{endo*},\ \mathbf{X}_{1:T,N-1}^{exo*}) = \mathbf{X}_{1:T,1}^{endo*}
\end{equation}

Now, we construct the convolution kernel \(K^{\prime}\) as follows:

\begin{equation}
K^{\prime} = (1-\alpha)K + \alpha S
\end{equation}

This ensures that the convolution operation in the Cross Only method matches the weighted sum in the Summation method. Specifically, the outputs are:

{\scriptsize
\begin{equation}
K^{\prime} * \mathrm{Stack}(\mathbf{X}_{1:T,1}^{endo*},\ \mathbf{X}_{1:T,N-1}^{exo*}) = (1-\alpha) \cdot (K * \mathrm{Stack}(\mathbf{X}_{1:T,1}^{endo*},\ \mathbf{X}_{1:T,N-1}^{exo*})) + \alpha \cdot \mathbf{X}_{1:T,1}^{endo*}
\end{equation}
}

This is precisely the expression used in the Summation method. Therefore, by choosing \(K^{\prime} = (1-\alpha)K + \alpha S\), we have proven that the Summation method and the Cross Only method are mathematically equivalent. The two methods yield the same result, differing only in their parameterization of the convolution operation.

However, experimental results indicate that treating endogenous and exogenous variables equally, as done in Cross Only, leads to poor performance. This may be attributed to the limited amount of data, which hinders the model's ability to learn the correct representation. These findings highlight the importance of distinguishing between endogenous and exogenous variables.

\section{Hyperparameter Sensitivity}
\label{sec: hyperparameter sensitivity}
\begin{figure}[!htbp]
\centering
\includegraphics[width=1\linewidth]{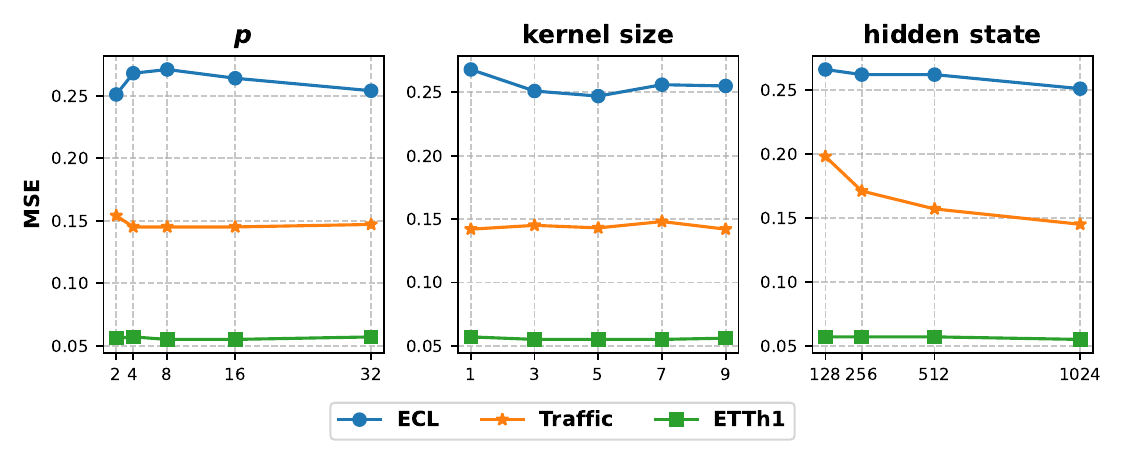}
\caption{Hyperparameter sensitivity with respect to patch size (\(p\)), convolution kernel size, and the number of hidden state units. Here, \(T=96\) and \(S=96\).}
\label{fig: Param Analysis Appendix}
\end{figure}

In this section, we evaluate the impact of three hyperparameters—patch size (\(p\)), convolution kernel size, and the number of hidden state units—on model performance (see Figure \ref{fig: Param Analysis Appendix}). The results reveal that the model's sensitivity to these parameters varies across different datasets. For the ETTh1 dataset, these hyperparameters have minimal effect on performance. However, the model's performance on the ECL and Traffic dataset is more sensitive to patch size and kernel size. Specifically, a larger kernel size leads to better performance on the ECL dataset, which may be attributed to the presence of a certain lead-lag effect. While, on the Traffic dataset, a larger number of hidden state units can result in better performance, likely due to a larger capacity of CrossLinear.

\section{Full Results of Model Generality}
\label{sec: full results}
To comprehensively demonstrate the effectiveness and versatility of cross-correlation embedding as a plug-and-play component across various time series forecasting models, we integrate it into five distinct architectures: SparseTSF, RLinear, PatchTST, DLinear, and Autoformer. By incorporating this embedding mechanism into diverse model structures, we aim to assess its generalizability and potential performance improvements across different forecasting paradigms. The complete experimental results, highlighting the impact of cross-correlation embedding on each model, are presented in Table \ref{table: full embedding generality}.

\section{Showcases}
\begin{figure}[!htbp]
\centering
\subfloat[CrossLinear \textcolor{red}{(Ours)}]{\includegraphics[width=0.5\linewidth]{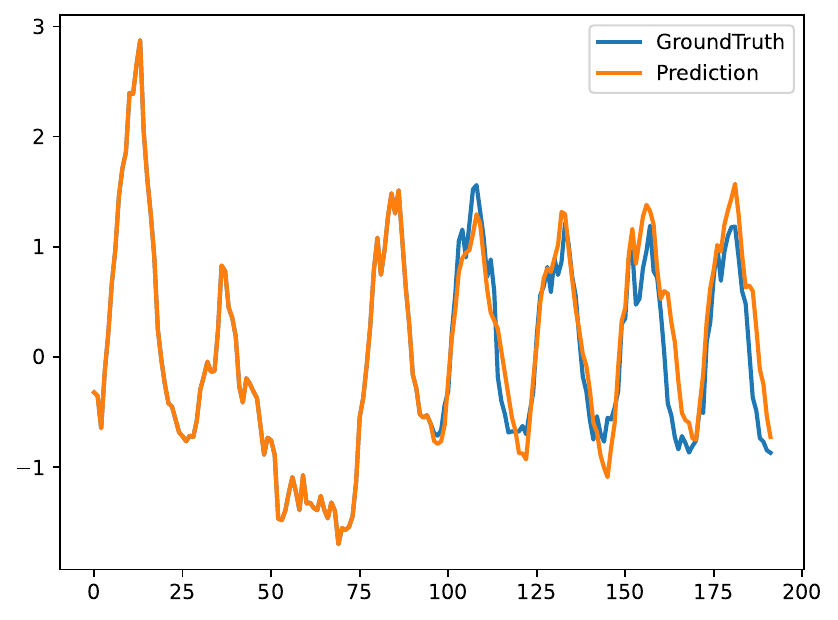}}
\subfloat[TimeXer]{\includegraphics[width=0.5\linewidth]{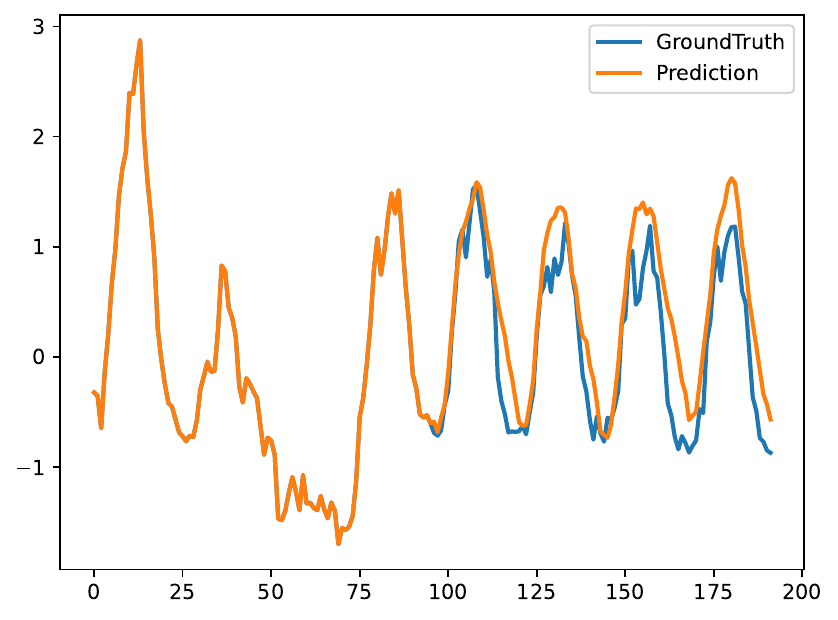}} \\
\subfloat[iTransformer]{\includegraphics[width=0.5\linewidth]{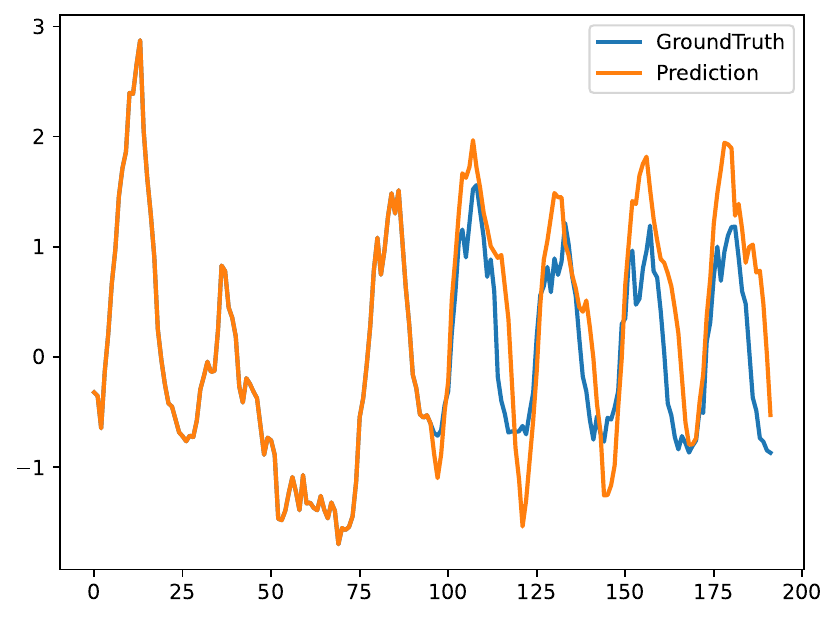}}
\subfloat[PatchTST]{\includegraphics[width=0.5\linewidth]{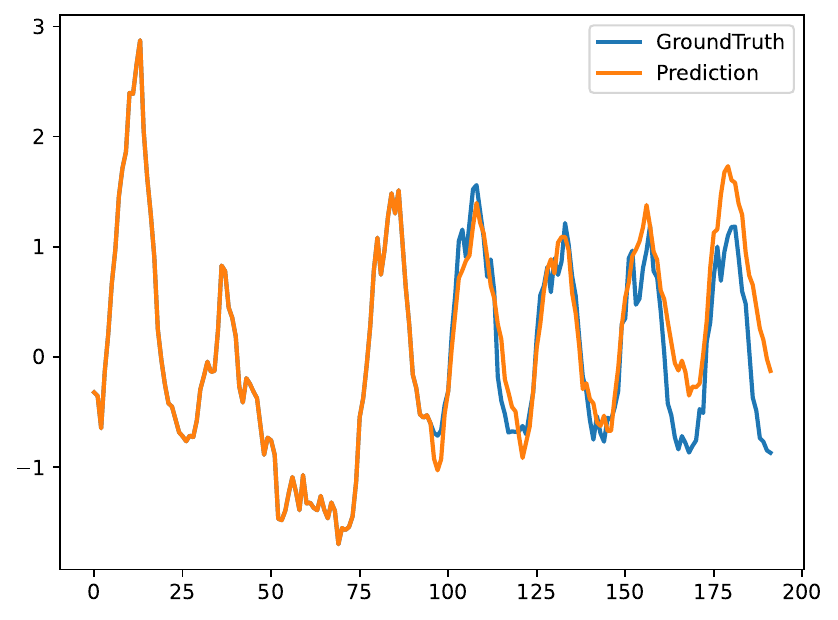}} \\
\subfloat[TiDE]{\includegraphics[width=0.5\linewidth]{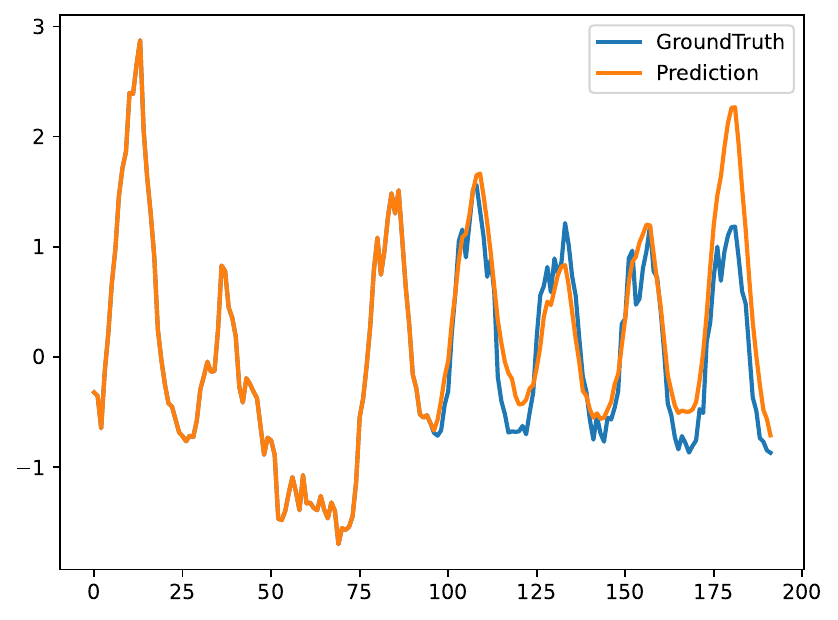}}
\subfloat[Autoformer]{\includegraphics[width=0.5\linewidth]{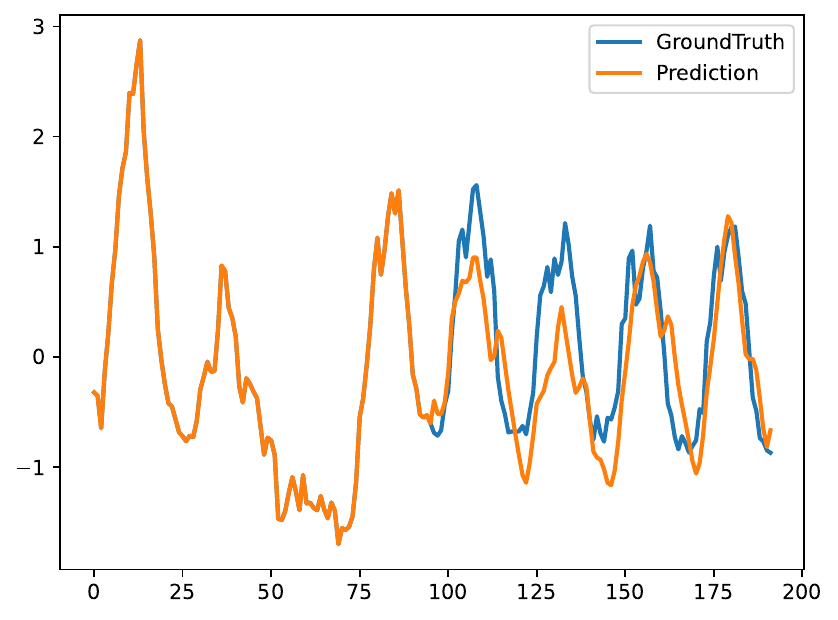}}
\caption{Visualization of results on the ECL dataset. Here, \(T=96\) and \(S=96\).}
\label{fig: Showcases}
\end{figure}

Here we provide showcases for visualization, as shown in Figure \ref{fig: Showcases}. We display the ground truth values and the forecasting results. We compare the forecasting performance of the proposed CrossLinear with five comparable baseline models, including TimeXer, iTransformer, PatchTST, TiDE, and Autoformer. Each model takes the input time series data with a length of 96 and performs forecasting tasks with a prediction horizon of 96.

\end{document}